\documentclass[10pt,twocolumn,letterpaper]{article}

\usepackage{cvpr}              

\usepackage[T1]{fontenc}
\usepackage{graphicx}
\usepackage{array}
\usepackage{multirow}
\usepackage{booktabs}

\usepackage{algorithm}
\usepackage{algorithmic}
\usepackage{enumitem}

\usepackage{amsmath}
\DeclareMathOperator*{\argmin}{arg\,min}
\usepackage{amssymb}
\usepackage[table,xcdraw]{xcolor}

\newcommand{\yi}[1]{\textcolor{black}{#1}}
\newcommand{\hazel}[1]{\textcolor{black}{#1}}
\newcommand{\jinglong}[1]{\textcolor{black}{#1}}

\definecolor{cvprblue}{rgb}{0.21,0.49,0.74}
\usepackage[pagebackref,breaklinks,colorlinks,allcolors=cvprblue]{hyperref}

\def\paperID{11235} 
\def\confName{CVPR}
\def\confYear{2025}

\begin{document}
\title{Continual Learning for Segment Anything Model Adaptation}

\author{
Jinglong Yang\textsuperscript{1,2,}\footnotemark[1], 
Yichen Wu\textsuperscript{2,}\footnotemark[1], 
Jun Cen\textsuperscript{3}, 
Wenjian Huang\textsuperscript{1}, 
Hong Wang\textsuperscript{4,}\footnotemark[2], 
Jianguo Zhang\textsuperscript{1,}\footnotemark[2] \\
\textsuperscript{1}Southern University of Science and Technology \quad
\textsuperscript{2}City University of Hong Kong \\
\textsuperscript{3}Hong Kong University of Science and Technology \quad
\textsuperscript{4}Tencent YouTu Lab \\
{\tt\small yangjl2022@mail.sustech.edu.cn},~~
{\tt\small wuyichen.am97@gmail.com}, ~~
{\tt\small jcenaa@connect.ust.hk} \\
{\tt\small wjhuang@pku.edu.cn}, ~~
{\tt\small hongwang9209@hotmail.com}, ~~
{\tt\small zhangjg@sustech.edu.cn}
} 
\maketitle
\renewcommand{\thefootnote}{\fnsymbol{footnote}}
\footnotetext[1]{Co-first author}
\renewcommand{\thefootnote}{\arabic{footnote}}
\renewcommand{\thefootnote}{\fnsymbol{footnote}}
\footnotetext[2]{Corresponding author}
\renewcommand{\thefootnote}{\arabic{footnote}}

\begin{abstract}

Although the current different types of SAM adaptation methods have achieved promising performance for various downstream tasks, such as prompt-based ones and adapter-based ones, most of them belong to the one-step adaptation paradigm. In real-world scenarios, we are generally confronted with the dynamic scenario where the data comes in a streaming manner. Driven by the practical need, in this paper, we first propose a novel Continual SAM adaptation (CoSAM) benchmark with 8 different task domains and carefully analyze the limitations of the existing SAM one-step adaptation methods in the continual segmentation scenario. Then we propose a novel simple-yet-effective Mixture of Domain Adapters (MoDA) algorithm which utilizes the Global Feature Tokens (GFT) and Global Assistant Tokens (GAT) modules to help the SAM encoder extract well-separated features for different task domains, and then provide the accurate task-specific information for continual learning.
Extensive experiments demonstrate that our proposed MoDA obviously surpasses the existing classic continual learning methods, as well as prompt-based and adapter-based approaches for continual segmentation. Moreover, after sequential learning on the CoSAM benchmark with diverse data distributions, our MoDA maintains highly competitive results in the natural image domain, approaching the zero-shot performance of the original SAM, demonstrating its superior capability in knowledge preservation. Notably, the proposed MoDA can be seamlessly integrated into various one-step adaptation methods of SAM, which can consistently bring obvious performance gains. Code is available at \href{https://github.com/yangjl1215/CoSAM}{https://github.com/yangjl1215/CoSAM}
\end{abstract}    
\section{Introduction}
\yi{Image segmentation is a foundational task in computer vision \cite{maskformer, oneformer}, essential for enabling machines to understand visual content across various applications. Recently, Segment Anything Model (SAM)~\cite{sam} has introduced a groundbreaking shift in this field, achieving remarkable versatility and accuracy. With its ability to perform high-quality segmentation across diverse settings, SAM represents a major leap forward, setting new benchmarks and significantly expanding the scope of automated segmentation.}

\begin{figure}[t]
	\centering
	\includegraphics[width=0.5\textwidth]{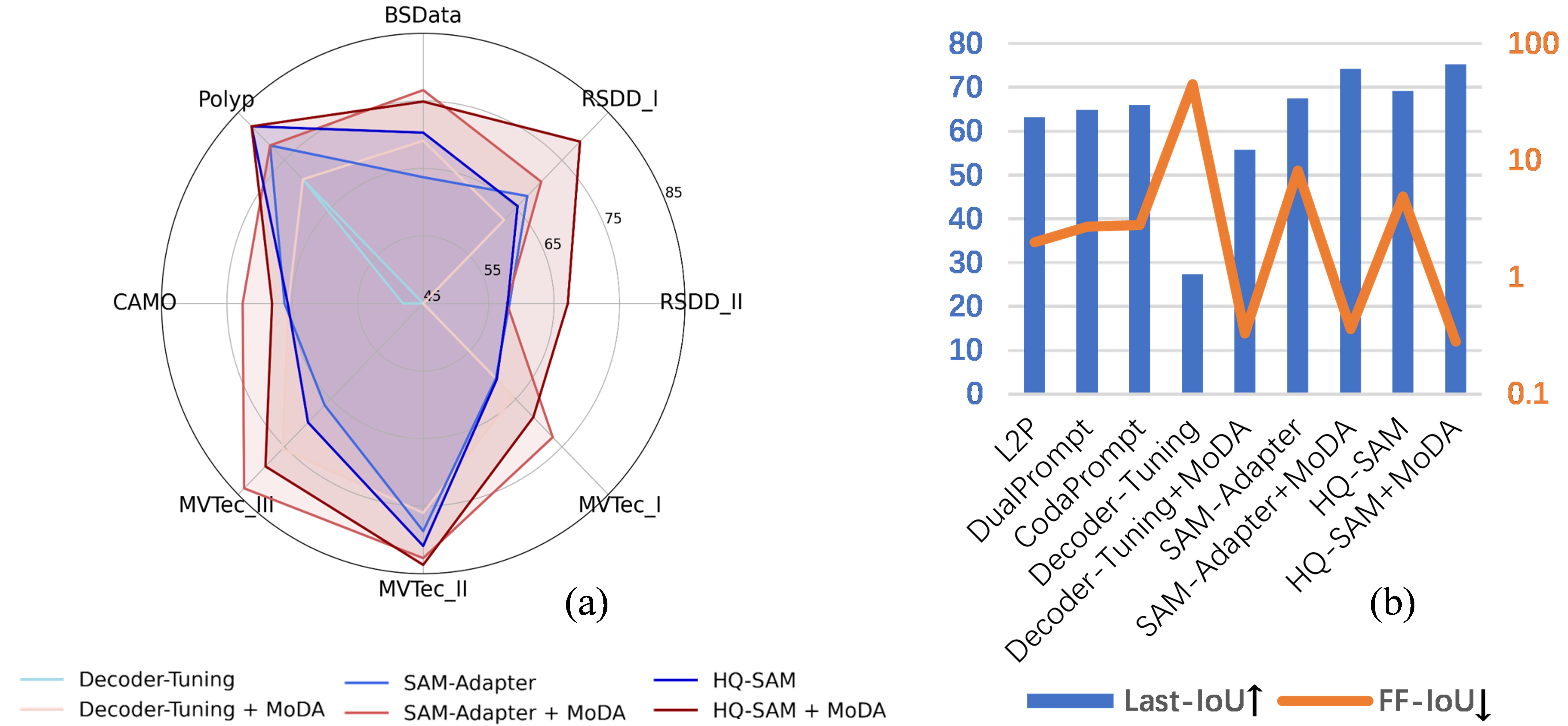}
    \caption{(a) Radar plot illustrating the performance of different segmentation methods across various tasks. Our method (+MoDA) significantly enhances the performance of the base model; (b) the average performance (i.e., Last-IoU) and average forgetting measure (i.e., FF-IoU) of different methods.  }
    \vspace{-5mm}
    \label{radar_chart}
\end{figure}

\yi{Although SAM provides a solid foundation for segmentation tasks, recent studies indicate that its performance is limited in certain specific sub-domains~\cite{sam_struggle_chengmingming, sam_not_perfect, sam_adapter}. To address this issue, various algorithms have been developed, each proposes different strategies to enhance SAM’s performance on downstream tasks. For instance, SAM-Adapter \cite{sam_adapter} integrates multiple learnable MLP blocks to enhance segmentation in complex scenarios like camouflage detection and polyp segmentation. Surgical-SAM \cite{surgical_sam} uses Low-Rank Adaptation (LoRA) \cite{lora} for increasing accuracy in robotic surgery. HQ-SAM \cite{sam_hq} and MSA~\cite{msa} employ adapters with diverse structures to tailor SAM for various granularities and modalities, demonstrating proactive efforts to improve its efficacy in specific domains.} 

\yi{However, these methods primarily rely on one-step adaptation, which is limited in real-world scenarios where models face continuous data streams and need ongoing updates. In such dynamic scenarios, one-step adaptation methods focus solely on the most recently finetuned task, making it difficult for models to handle new tasks while preserving performance on previous ones. 
Therefore, how to utilize the capabilities of SAM across a sequence of continual tasks to quickly adapt to new tasks while maintaining the performance of old tasks is a challenging problem. To investigate SAM's potential in the continual scenario, we construct a benchmark for continual segmentation, called Continual SAM Adaptation Benchmark (CoSAM), which aims to systematically evaluate SAM-related algorithms's performance within the streaming scenarios. Specifically, CoSAM offers a set of 8 tasks covering diverse domains, including industrial defects, medical imaging, and camouflaged objects, to serve as a realistic and effective benchmark for evaluating current methods.}

\yi{Leveraging the constructed CoSAM, we select six representative continual learning (CL) algorithms, modify and integrate them into SAM, and evaluate their performance.  Specifically, these CL algorithms include classic methods such as LwF \cite{lwf}, EWC \cite{ewc}, and ER \cite{er}, along with recent, more competitive prompt-based methods like L2P \cite{l2p}, Dual-Prompt \cite{dualprompt}, and Coda-Prompt \cite{codaprompt}. These prompt-based methods typically construct a prompt pool containing key-value pairs of tokens. By matching the image features extracted by pre-trained model with the well-learned key tokens, the model can utilize the corresponding value tokens as the task-specific information to boost performance.} \hazel{However, we find that opposite to the general performance trend for classification tasks with foundation models, prompt-based CL methods are inferior to classic
CL methods on the benchmark for continual segmentation tasks.}
\hazel{Through analyzing the reasons behind the suboptimal performance of leading prompt-based CL classification methods on our proposed continual segmentation benchmark, CoSAM, we identified two key factors: 1) Unlike ImageNet-pretained encoder widely adopted by prompt-based methods for helping extract well-separated key tokens across different tasks in the classification scenario, the SAM encoder is designed to interact with user prompts and it cannot naturally generate task-distinguishing key tokens (see Fig.~\ref{fig-tsne}), which makes it difficult for prompt-based methods to capture high-level semantic information for accurately identifying different task domains in the segmentation scene; 2) Compared to the coarse-grained classification task, the segmentation task focuses on the pixel-wise prediction with finer granularity. For prompt-based methods, simply concatenating the value tokens to adjust high-level feature spaces is not enough for handling this delicate segmentation task~\cite{prompt-seg}.}


\hazel{Motivated by the aforementioned analysis, we carefully propose a Mixture of Domain Adapters (MoDA) algorithm, which can be easily integrated into the existing SAM one-adaptation methods for more accurate continual segmentation.} \yi{Specifically, we introduce the Global Feature Tokens (GFT) and Global Assistant Tokens (GAT) modules, which enable the pre-trained segmentation encoder to capture high-level global features for better distinguishing different data domains. By interacting with image tokens at each layer of the transformer within the frozen SAM network, the model can extract task-specific information that represents the global features of the input image (see Fig.~\ref{fig-tsne}(c)).} \hazel{With the guidance of the optimized GFT, our model can more accurately query the task-specific information, which largely alleviates catastrophic forgetting and makes itself able to automatically switch between the original SAM and the adapter-based SAM depending on the incoming data domain during the inference phase. To the best of our knowledge, we are the first to deeply explore the potential of SAM in continual segmentation.}
To sum up,  our contributions can be summarized as follows:
 \begin{itemize}[leftmargin=3.5mm, itemsep=0mm, topsep=0.5 mm]
 \vspace{1mm}
    \item[$\circ$] \hazel{We establish the new CoSAM benchmark consisting of 8 tasks with quite different segmentation scenarios and provide the corresponding evaluation metrics to assess the performance of SAM-based CL algorithms, providing a key platform for future research on CL effectiveness in segmentation-based foundational models.}
      \vspace{1mm}
    \item[$\circ$] \hazel{We carefully modify the current top-performing CL methods to make them applicable to the benchmark and deeply analyze why they struggle on CoSAM. Building on these insights, we propose a simple-yet-effective MoDA algorithm, which provides more accurate task classification information (see Fig.~\ref{fig-tsne} and Table~\ref{tab:num-GAT}). It can be easily integrated into different SAM one-step adapation methods for better segmentation performance and lower forgetting rate (see Fig.~\ref{radar_chart} and Table~\ref{tab:ablat}).}
    \vspace{1mm}
    \item[$\circ$] \yi{We conduct comprehensive experiments and ablation studies to demonstrate the effectiveness of the proposed MoDA,} \hazel{which obviously outperforms the existing classic CL methods, prompt-based, and adapter-based ones on the CoSAM benchmark (see Fig.~\ref{radar_chart}) as well as on the natural image domain(see Table~\ref{tab:natural_domain}), demonstrating a stronger capability in knowledge preservation.}
\end{itemize}

\section{Related Work}
\jinglong{\textbf{Segment Anything Model.} SAM~\cite{sam} showcases remarkable versatility and accuracy in image segmentation. However, Some studies~\cite{sam_not_perfect, sam_struggle_chengmingming, sam_adapter, msa} have identified limitations of SAM in certain domains, such as industrial defect detection, medical imaging, and camouflage animals. These findings have spurred a range of research efforts~\cite{sam_hq, sam_adapter, surgical_sam, masam, msa, adaptivesam} that explore parameter-efficient tuning methods to enhance SAM’s adaptability in specific domains.}
\yi{For the continual segmentation benchmark, UnCLe SAM~\cite{unclesam} explores the combination of CL with SAM. However,} rather than bringing CL methodologies into SAM adaptation, it introduces SAM into traditional CL segmentation tasks, which lack interactive inputs like bounding boxes. Therefore, UnCLe SAM needs to deploy an external ResNet~\cite{resnet} to generate bounding boxes in the CL setting. These bounding boxes are then fed into a frozen SAM model to produce segmentation masks, making it closer to a continual learning approach for object detection rather than SAM adaptation. In contrast, our approach brings CL into SAM adaptation, enhancing SAM’s ability to handle continuously evolving domains. 

\textbf{Parameter Efficient Tuning (PEFT).} The computer vision field has actively explored PET, demonstrating its effectiveness through various approaches. \cite{li2022exploring} showed that fine-tuning the Vision Transformer for object detection requires minimal changes. \cite{liu2023explicit} enhanced adaptability with Explicit Visual Prompting, integrating visual cues into the Adapter. Prompt Tuning methods \cite{vpt, lester2021power, li2021prefix}, involving the addition of learnable tokens to image patches, have significantly improved transfer learning with minimal architectural modifications.


\jinglong{\textbf{Continual Learning.} Continual Learning aims to enable models to learn new tasks without forgetting previously learned knowledge. Traditional CL methods include regularization-based approaches \cite{lwf, ewc, si, rwalk}, which primarily use knowledge distillation to retain prior information during new task training. Another popular approach is replay-based methods \cite{er, icarl, dgr, der}, which retain a small subset of previous data to periodically refresh old knowledge as new tasks are learned.}

\jinglong{As foundational models gain prominence in the deep learning field, prompt-based CL methods \cite{l2p, dualprompt, codaprompt, cprompt} have emerged to leverage the power of large pre-trained models. These approaches mitigate catastrophic forgetting by utilizing a pre-trained encoder to extract image features and query a prompt pool. The prompts retrieved from this pool offer task-specific context, enhancing the model's ability to make task-aware predictions.}

\jinglong{Among these methods, Learning to Prompt (L2P) \cite{l2p} pioneered the use of a prompt pool, in which, during training, the value tokens within the pool are optimized to adapt to the current task while the key tokens are pulled closer to the input embeddings. This allows accurate querying during inference, enabling the input image to retrieve the most suitable value token from the prompt pool. Building upon this, DualPrompt \cite{dualprompt} addressed L2P's limitation of not explicitly separating task-specific prompts from task-sharing prompts. Task-sharing prompts facilitate positive knowledge transfer across different tasks. CodaPrompt \cite{codaprompt} further refines the prompt pool query mechanism by transitioning from hard selection to a soft selection process. Leveraging attention mechanisms, it calculates the similarity between input embeddings and key tokens in the prompt pool to determine the weighting of each value token.}

\section{Continual SAM Adaptation Benchmark}
\yi{Despite the growing number of SAM-based adaptation methods, most focus on enhancing SAM’s performance for a single downstream task, using a one-step adaptation approach \cite{sam_adapter, sam_hq, decoder-tuning}. This overlooks the demands of real-world scenarios that require continual adaptation for ongoing segmentation tasks.~Therefore, we first propose the Continual SAM Adaptation Benchmark (CoSAM) to measure the performance of the SAM adaptation algorithm in the CL setting. In this section, we provide a detailed description of the constructed CoSAM benchmark.}

\begin{figure*}[!t]
    \centering
    \includegraphics[width=0.9\textwidth]{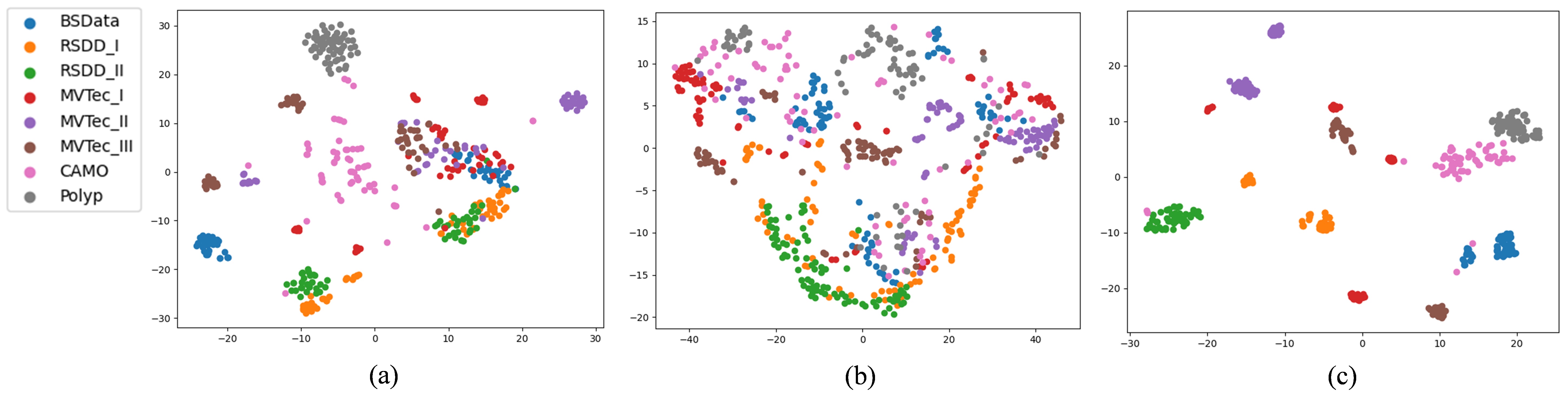}
    \vspace{-3mm}
    \caption{T-SNE visualization of features of different datasets extracted by (a) ImageNet Pre-trained Encoder, (b) SAM Encoder and (c) Global Feature Token. We use mean operation to reduce the feature dimension from $\mathbb{R}^{L \times C}$ to $\mathbb{R}^{1 \times C}$}
    \label{fig-tsne}
    \vspace{-4mm}
\end{figure*}

\textbf{\jinglong{Dataset.}}~\yi{To make our constructed dataset more representative of real-world scenarios, we select segmentation datasets from diverse domains, moving beyond the traditional focus on classification-based CL tasks, which are often confined to similar domains like CIFAR-100~\cite{cifar100} and ImageNet-R~\cite{imagenetr}. The CoSAM benchmark, as shown in Table~\ref{tab:datasets}, comprises five datasets organized into eight tasks. Specifically, we divided two of the datasets into multiple tasks: the RSDDs dataset~\cite{RSDD}, which includes two distinct defect types, was split into two tasks, while the MVTec dataset~\cite{mvtec}, containing 15 types of common objects with relatively small data samples, was grouped into three tasks to mitigate severe data imbalance. These eight tasks, spanning diverse domains, provide a comprehensive setting to assess SAM's continual adaptation capabilities. For detailed descriptions, please refer to the supplementary material.}


\begin{table}[t]
    \centering
    \caption{Datasets used in the CoSAM benchmark. The \textit{Size} column indicates the number of samples in each dataset, the \textit{Tasks} column specifies how many tasks each dataset contributes to the benchmark, and the \textit{Content} column describes the specific segmentation tasks addressed by each dataset.}
    \vspace{-2mm}
    \begin{tabular}{l c c p{3.5cm}} 
        \toprule
        \footnotesize \textbf{Dataset} & \footnotesize \textbf{Size} & \footnotesize \textbf{Tasks} & \qquad \quad ~~ \footnotesize \textbf{Content} \\
        \midrule   
        BSData \cite{BSData} & 488 & 1 & Industrial surface defects \\
        \hline
        RSDD \cite{RSDD} & 318 & 2 & Defects of high-speed track \& heavy-duty track\\
        \hline
        MVTec \cite{mvtec} & 676 & 3 & Various defects in 15 common objects \\
        \hline
        CAMO \cite{camo} & 1250 & 1 & Camouflaged object detection \\
        \hline
        Polyp \cite{polyp} & 1000 & 1 & Medical polyp detection \\
        \bottomrule
    \end{tabular}
    \vspace{-4mm}
    \label{tab:datasets}
\end{table}
\yi{\textbf{Evaluation Metrics.} To examine the final average accuracy, we utilize the metrics Last-IoU and Last-BIoU to assess the segmentation performance of all tasks after learning all sequential tasks. For clarity, we first define the average performance after training $t$ tasks, denoted as $\text{IoU}_t$, }
\begin{equation}
    \text{IoU}_t = \frac{1}{t} \sum_{k=1}^t \text{IoU}_{k,t},
    \quad \text{BIoU}_t = \frac{1}{t} \sum_{k=1}^t \text{BIoU}_{k,t},
\end{equation}
\yi{where $\text{IoU}_{k,t}$ and $\text{BIoU}_{k,t}$ represent the IoU/BIoU evaluated on the test set of the $k$-th task after training on $t$ tasks. Then, the Last-IoU and Last-BIoU can be defined as,}
\begin{equation}
    \text{Last-IoU} = \text{IoU}_N , \quad
    \text{Last-BIoU}= \text{BIoU}_N.
\end{equation}
\yi{To show the average segmentation results over the training trajectory during sequential training, we also adopt the Avg-IoU and Avg-BIoU metrics, as detailed below:}
\begin{equation}
     \text{Avg-IoU} = \frac{1}{N} \sum_{t=1}^N \text{IoU}_t, \quad \text{Avg-BIoU} = \frac{1}{N} \sum_{t=1}^N \text{BIoU}_t.
\end{equation}
\yi{To measure forgetting performance, we first define $f_{k,t}$ as the forgetting on task $t$ after trained on all $t$ tasks, }
\begin{equation}
        f_{k,t} = \max_{j \in \{1,\cdots,t-1\}} \text{IoU}_{k,j} - \text{IoU}_{k,t}.
\end{equation}
\yi{Then, we can get the average forgetting measure after training on all $N$ tasks, i.e., FF-IoU,}
\begin{equation}
    \text{FF-IoU} = \frac{1}{N\!-\!1} \sum_{k=1}^{N\!-\!1} f_{k,N}.
\end{equation}
And the metric FF-BIoU is defined in a similar way.

\section{Method}\label{sec:method}
\subsection{Preliminary}
\yi{Let $\mathcal{D}_{tr}=\{D^1_{tr},\dots, D^N_{tr}\}$ and $\mathcal{D}_{te}=\{D^1_{te},\dots, D^N_{te}\}$ denote the sequential training \jinglong{datasets} and testing \jinglong{datasets}, respectively. Here, the notation $D_{tr}^i$ and $D_{te}^i$ refer to the training/testing dataset from the $i$-th task, and $N$ represents the total number of tasks. The $f_\theta(\cdot)$ denotes the SAM parametrized with $\theta$, $\theta'$ specifically refer to the weights of SAM encoder, and $P_{img}\in\mathbb{R}^{L\times C}$ refers to the image patch embedding, \jinglong{where $L$ is the length of image patch token sequence and $C$ is the number of dimensions of each token, i.e., 768 for the ViT-b model.} Let $\mathcal{P}=\{K,\Phi\}$ represent the adapter pool, where the adapter $\Phi=\{\phi_1,...,\phi_N\}$, the learnable key $K=\{k_1, ..., k_N\}$.
For clarity, in this paper, we define the $t$-th task as the current training task, and then $\{D_{tr}^1,..., D_{tr}^{t-1}\}$ are the previously trained datasets.}

\begin{figure*}[!t]
	\centering
	\includegraphics[width=\textwidth]{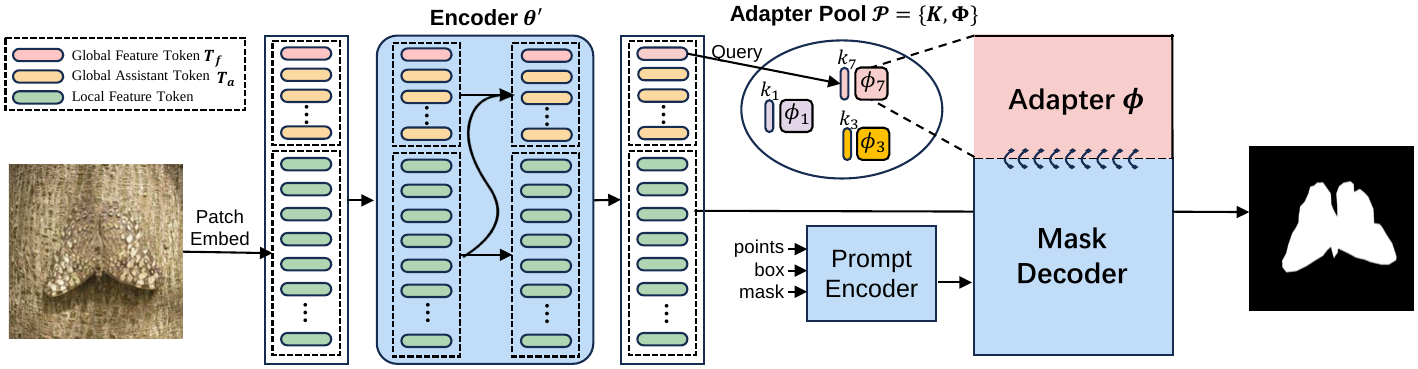}
        \vspace{-4mm}
	\caption{Illustration of our MoDA, including the global feature token $T_f$, global assistant token $T_a$, and the adapter $\phi$. Based on $T_f$ and $T_a$, we generate enhanced queries that capture global discriminative features, which help select the corresponding task-specific adapter $\phi$.}
    \vspace{-4mm}
        \label{fig-architect}
\end{figure*}

\subsection{Mixture of Domain Adapters (MoDA)}\label{sec:fail}
\yi{Using the constructed CoSAM benchmark, we select six representative CL algorithms, including classic methods like LWF, EWC, and ER, as well as recent, competitive prompt-based approaches such as L2P, Dual-Prompt, and Coda-Prompt, which leverage the power of foundation models. Interestingly, although prompt-based methods generally excel in classification CL tasks, their performance on the CoSAM benchmark drops significantly, even falling below that of the classic methods, as shown in Table \ref{tab:combined_performance_comparison}.}

\yi{Examining these prompt-based CL methods reveals that they generally build a prompt pool with key-value token pairs to store task-specific knowledge during sequential training. During testing, the model matches the image features extracted by the pre-trained encoder with the learned key tokens in the pool, allowing it to retrieve the corresponding task-specific knowledge (i.e., paired value tokens) to enhance performance. However, simply adapting these prompt-based methods from classification to segmentation tasks presents two main challenges, leading to a substantial drop in performance, as outlined below:}
\begin{itemize}[leftmargin=3mm, itemsep=0mm, topsep=-0.5 mm]
    \item[$\circ$] \yi{\textit{Key Matching Mechanism Fai›lure.}  In classification tasks, prompt-based methods capitalize on the ImageNet-pre-trained encoder’s ability to extract high-level global features, leading to well-separated key tokens across tasks. However, when directly applied to segmentation tasks, the SAM encoder—designed to respond to user inputs like points or bounding boxes and emphasize local information over global features—does not naturally generate task-distinguishing key tokens. Without specific adaptations, the key matching mechanism fails to effectively differentiate between tasks, as shown in Figure~\ref{fig-tsne}.}
    \item[$\circ$]  \yi{\textit{Lack of Accurate Feature Modulation.} Original prompt-based CL methods rely on choosing and concatenating value tokens to adjust high-level feature spaces, which is adequate for influencing classification decisions. However, segmentation tasks demand spatially detailed feature transformations to accurately delineate object boundaries and regions, ensuring accurate pixel-wise predictions. This limitation hampers the effectiveness of prompt-based methods when directly applied to segmentation.}
\end{itemize}

\yi{To address these two challenges, we propose the Mixture of Domain Adapters (MoDA) algorithm, which is explained in detail in the following two subsections. The overall framework is illustrated in Fig.~\ref{fig-architect}. }



\subsubsection{Global Feature Token \& Global Assistant Token}
\yi{Against the first problem, we introduce the Global Feature Token (GFT), denoted as $T_f \in \mathbb{R}^{1 \times C}$, designed to assist the SAM encoder in capturing high-level global features essential for distinguishing between tasks. However, a single GFT alone lacks the flexibility to capture task-specific information effectively. To overcome this limitation, we introduce a set of Global Assistant Tokens (GATs), denoted as $T_a \in \mathbb{R}^{N_a \times C}$, which interact with image tokens across each layer of the transformer. This interaction allows the model to progressively extract global, task-specific features from the input image. } 

\yi{\!\!As shown in Eqn.~(\ref{eqn:query}), during training, we concatenate the global token \jinglong{$T_f$ and the global assistant tokens $T_a$} with the image patch embeddings $P_{img}$. These concatenated inputs are passed through the SAM encoder $f_{\theta'}$ to generate the query $q$, representing the task-specific global information. }
\begin{equation}
\label{eqn:query}
   q = f_{\theta'}(T_f,T_a,P_{img})\left[0\right].
\end{equation}
\yi{Then, we update the global token $T_f$, the global assistant tokens $T_a$, and the learnable key $k$ for matching by calculating the following cross-entropy loss:}
\begin{equation}
\label{eqn:ce}
    \mathcal{L}_{ce} = -\log \frac{\exp s_{gt}}{\sum_{i=1}^{t} \exp s_i}, \quad s_i= cos(q, k_i),
\end{equation}
\jinglong{where $cos(\cdot)$ denotes the cosine similarity and $s_{gt}$ represents the similarity score of the ground truth task. }

\yi{By incorporating the GFT $T_f$ and GAT $T_a$ specifically designed for SAM, we enable seamless integration within SAM's internal structure. This design contrasts with prior approaches that applied SAM to CL, which either suffered from poor compatibility or required additional structures to predict task IDs. Our approach eliminates these issues, providing a more effective solution for CL within SAM.}


\begin{table*}[!t]
\centering
\caption{Comparison of methods based on IoU and BIoU metrics, with results from five seeds (mean and standard deviation).}
\resizebox{0.9\textwidth}{!}{
\setlength{\tabcolsep}{7pt} 
\renewcommand{\arraystretch}{1.05}
\begin{tabular}{lccccccc}
\toprule

\multirow{2}{*}{\textbf{Method}} & \multicolumn{3}{c}{\textbf{IoU}} & \multicolumn{3}{c}{\textbf{BIoU}} \\ 
\cmidrule(lr){2-4} \cmidrule(lr){5-7}
& \textbf{Last-IoU}$\uparrow$ & \textbf{Avg-IoU}$\uparrow$ & \textbf{FF-IoU}$\downarrow$ 
& \textbf{Last-BIoU}$\uparrow$ & \textbf{Avg-BIoU}$\uparrow$ & \textbf{FF-BIoU}$\downarrow$ \\
\midrule
Joint Training (Upper Bound) & 76.3\scriptsize{$\pm$0.31} & - & - & 66.9\scriptsize{$\pm$0.26} & - & - \\
SAM & 69.0 & - & - & 58.9 & - & - \\
\cline{1-7} 
LwF & 73.3\scriptsize{$\pm$0.72} & 72.0\scriptsize{$\pm$0.68} & 2.0\scriptsize{$\pm$0.39} 
& 63.7\scriptsize{$\pm$0.78} & 69.1\scriptsize{$\pm$0.69} & 2.3\scriptsize{$\pm$0.31} \\
EWC & 72.9\scriptsize{$\pm$1.25} & 72.3\scriptsize{$\pm$0.97} & 3.5\scriptsize{$\pm$0.56} 
& 63.2\scriptsize{$\pm$1.01} & 68.9\scriptsize{$\pm$0.82} & 3.9\scriptsize{$\pm$0.48} \\
ER & 73.0\scriptsize{$\pm$0.93} & 73.2\scriptsize{$\pm$0.71} & 2.9\scriptsize{$\pm$0.39} 
& 62.7\scriptsize{$\pm$0.89} & 69.8\scriptsize{$\pm$0.63} & 3.4\scriptsize{$\pm$0.30} \\
\cline{1-7}
L2P & 63.1\scriptsize{$\pm$1.15} & 56.5\scriptsize{$\pm$1.49} & 2.0\scriptsize{$\pm$0.43} 
& 52.5\scriptsize{$\pm$1.24} & 53.2\scriptsize{$\pm$1.31} & 3.3\scriptsize{$\pm$0.56} \\
Dual-Prompt & 64.9\scriptsize{$\pm$1.43} & 64.5\scriptsize{$\pm$1.63} & 2.7\scriptsize{$\pm$0.40} 
& 55.0\scriptsize{$\pm$1.71} & 61.1\scriptsize{$\pm$1.56} & 2.6\scriptsize{$\pm$0.39} \\
Coda-Prompt & 65.9\scriptsize{$\pm$1.45} & 63.8\scriptsize{$\pm$1.62} & 2.8\scriptsize{$\pm$0.37} 
& 55.3\scriptsize{$\pm$1.28} & 60.3\scriptsize{$\pm$1.43} & 2.1\scriptsize{$\pm$0.35} \\
\cline{1-7}
Decoder-Tuning & 27.3\scriptsize{$\pm$1.69} & 46.9\scriptsize{$\pm$1.37} & 45.0\scriptsize{$\pm$1.82} 
& 19.4\scriptsize{$\pm$1.57} & 43.3\scriptsize{$\pm$1.48} & 43.7\scriptsize{$\pm$1.66} \\
\rowcolor{orange!10} Decoder-Tuning + MoDA & 55.8{\scriptsize$\pm$0.58} (\textcolor{red}{$\uparrow 28.5$}) & 51.2{\scriptsize$\pm$0.52} (\textcolor{red}{$\uparrow 4.3$}) & 0.33{\scriptsize$\pm$0.12} (\textcolor{red}{$\downarrow 44.7$}) 
& 44.8{\scriptsize$\pm$0.49} (\textcolor{red}{$\uparrow 25.4$}) & 47.4{\scriptsize$\pm$0.43} (\textcolor{red}{$\uparrow 4.1$}) & 0.27{\scriptsize$\pm$0.09} (\textcolor{red}{$\downarrow 43.4$}) \\
SAM-Adapter & 67.4\scriptsize{$\pm$0.75} & 69.3\scriptsize{$\pm$0.63} & 8.2\scriptsize{$\pm$0.59} 
& 56.9\scriptsize{$\pm$0.81} & 65.8\scriptsize{$\pm$0.74} & 9.1\scriptsize{$\pm$0.65} \\
\rowcolor{orange!10} SAM-Adapter + MoDA & 74.3{\scriptsize $\pm$0.37} (\textcolor{red}{$\uparrow 6.9$}) & 73.4{\scriptsize $\pm$0.45} (\textcolor{red}{$\uparrow 4.1$}) & 0.36{\scriptsize $\pm$0.1} (\textcolor{red}{$\downarrow 7.8$}) 
& 64.3{\scriptsize $\pm$0.41} (\textcolor{red}{$\uparrow 7.4$}) & 69.9{\scriptsize$\pm$0.38} (\textcolor{red}{$\uparrow 4.1$}) & 0.27{\scriptsize$\pm$0.11} (\textcolor{red}{$\downarrow 8.8$})\\
HQ-SAM & 68.7\scriptsize{$\pm$0.59} & 70.9\scriptsize{$\pm$0.37} & 4.9\scriptsize{$\pm$0.44} 
& 58.6\scriptsize{$\pm$0.23} & 67.9\scriptsize{$\pm$0.35} & 9.3\scriptsize{$\pm$0.40} \\
\rowcolor{orange!10} HQ-SAM + MoDA & 75.2{\scriptsize$\pm$0.42} (\textcolor{red}{$\uparrow 6.5$}) & 74.9{\scriptsize$\pm$0.61} (\textcolor{red}{$\uparrow$ 4.0}) & 0.28{\scriptsize$\pm$0.09} (\textcolor{red}{$\downarrow 4.6$})
& 65.3{\scriptsize$\pm$0.37} (\textcolor{red}{$\uparrow 6.7$}) & 71.6{\scriptsize$\pm$0.50} (\textcolor{red}{$\uparrow 3.7$}) & 0.14{\scriptsize $\pm$0.07} (\textcolor{red}{$\downarrow 9.2$})\\

\bottomrule
\end{tabular}
}
\label{tab:combined_performance_comparison}
\end{table*}

\begin{table}[!t]
\centering
\caption{Performance on the Natural Domain, with mean and standard deviation, reflecting the impact on the original SAM's ability.}
\resizebox{0.48\textwidth}{!}{%
\setlength{\tabcolsep}{5pt} 
\renewcommand{\arraystretch}{1.05}
\begin{tabular}{lccc}
\hline
\textbf{Method} & \textbf{IoU}$\uparrow$ & \textbf{BIoU}$\uparrow$  \\
\hline
Joint Training &  76.9\scriptsize{$\pm$0.51}  & 65.4\scriptsize{$\pm$0.68}\\
SAM & \textbf{78.1} & \textbf{67.8} \\
\cline{1-3} 
LwF &  72.5\scriptsize{$\pm$0.74} & 59.3\scriptsize{$\pm$0.49}\\
EWC &  72.4\scriptsize{$\pm$0.53} & 60.8\scriptsize{$\pm$0.41}\\
ER & 75.9\scriptsize{$\pm$0.72} & 64.7\scriptsize{$\pm$0.55}\\
\cline{1-3}
L2P & 67.8\scriptsize{$\pm$1.27} & 56.2\scriptsize{$\pm$0.97}\\
Dual-Prompt & 66.5\scriptsize{$\pm$1.54} & 54.4\scriptsize{$\pm$1.32}\\
Coda-Prompt & 69.9\scriptsize{$\pm$1.77} & 57.5\scriptsize{$\pm$1.46}\\
\cline{1-3}
Decoder-Tuning & 0  & 0\\
\rowcolor{orange!10} Decoder-Tuning + MoDA & 76.9 {\scriptsize$\pm$0.12} (\textcolor{red}{$\uparrow 76.9$}) & 66.8{\scriptsize$\pm$0.10} (\textcolor{red}{$\uparrow 66.8$})\\
SAM-Adapter & 62.7\scriptsize{$\pm$0.54} & 53.1 \scriptsize{$\pm$0.46}\\
\rowcolor{orange!10}  SAM-Adapter + MoDA & 77.5{\scriptsize$\pm$0.63} (\textcolor{red}{$\uparrow 14.8$}) & 67.2{\scriptsize $\pm$0.57} (\textcolor{red}{$\uparrow 14.1$})\\
HQ-SAM & 72.7\scriptsize{$\pm$0.68} & 61.2\scriptsize{$\pm$0.53}\\
\rowcolor{orange!10}  HQ-SAM + MoDA &  77.6{\scriptsize$\pm$0.83} (\textcolor{red}{$\uparrow 4.9$}) & 67.0{\scriptsize$\pm$0.62} (\textcolor{red}{$\uparrow 5.8$})\\

\hline
\end{tabular}%
}
\vspace{-4mm}
\label{tab:natural_domain}
\end{table}

\subsubsection{Task Specific Adapter}
\yi{For each task $D_i, i=1,...,N$, we assign a unique adapter $\phi_i$ which is fine-tuned on $D_i$ to capture the task-specific information. Specifically, the adapter is trained to minimize the following loss $\mathcal{L}_{\text{mask}}$: }
\begin{equation}
\label{eqn:mask_loss}
    \mathcal{L}_{\text{mask}} = \mathcal{L}_{\text{Dice}}(f_{\theta}(x,\phi_t), y) + \mathcal{L}_{ce}(f_{\theta}(x,\phi_t), y),
\end{equation}
\yi{where $f_\theta$ is the SAM model, $\mathcal{L}_{\text{Dice}}$ refer to Dice loss and $\mathcal{L}_{ce}$ is the cross-entropy loss. }





\begin{algorithm}[!t]
    \caption{Mixture of Domain Adapters (MoDA) }\label{alg:moda}
    \begin{algorithmic}[1]
        \REQUIRE{Pre-trained SAM $\theta$, Tasks $\mathcal{D}=\{D_1,\dots,D_N\}$}
        \ENSURE{Adapter Pool $\mathcal{P}=\{K:\Phi\}$, global feature token $T_f \in \mathbb{R}^{1 \times C}$ and global assistant token $T_a \in \mathbb{R}^{N_a \times C}$}
        \STATE Initialize $\mathcal{P}=\{K:\Phi\}$, $T_f, T_a$.

        \FOR {$D_t \in \{D_1, \dots, D_N \}$}
            \FOR{$\{x, y\} \in D_t$}
                \STATE{Update $\phi_t$ with $\mathcal{L}_{\text{mask}}$ in Eqn.~(\ref{eqn:mask_loss}).}
            \ENDFOR
            \STATE Select exemplars for Memory Bank $\mathcal{M}$.
            \FOR{$x \in \mathcal{M}$}
                \STATE $q = f_{\theta'}(T_f,T_a,P_{img})\left[0\right]$
                \STATE Update $T_f, T_a, K$ with $\mathcal{L}_{ce}$ in Eqn.~(\ref{eqn:ce}).
            \ENDFOR
        \ENDFOR
    \end{algorithmic}
\end{algorithm}
\yi{The complete optimization procedure is detailed in Alg.~\ref{alg:moda}. During inference, our approach first extracts the global image feature $q$, as shown in Eqn.~(\ref{eqn:query}). This query $q$ is then compared with each key $k_i$ in the adapter pool $\mathcal{P}$ (i.e., $\arg\max_{i} cos(q,k_i)$) to identify the closest matching adapter $\phi_i$. By enhancing the effectiveness of task ID identification and using adapters to store task-specific knowledge, our approach significantly improves performance on continual learning segmentation tasks. }  







\section{Experiments}\label{sec:exps}
\subsection{Experimental Details}
\jinglong{Our proposed MoDA is implemented based on PyTorch on eight NVIDIA Tesla 3090 GPUs. The learning rate $\eta$ is set to 0.001. Batchsize is set to 4. Each experiment runs 24 epoches with Adam optimizer. The numbers of Global Feature Tokens (GFT) and Global Assistant Tokens (GAT), \emph{i.e.}, $N_{f}$ and $N_{a}$, are set to 1 and 10, respectively. The memory bank contains 10 samples per task. We repeat all experiments five times and present the average values and standard deviations in the paper.
}
\subsection{Comparison Methods}
\hazel{Currently, the existing representative CL methods are mainly developed for class-incremental learning scenarios, including classic CL approaches: LwF~\cite{lwf}, EWC~\cite{ewc}, ER~\cite{er}, and prompt-based CL ones: L2P~\cite{l2p}, Dual-Prompt~\cite{dualprompt}, and Coda-Prompt~\cite{codaprompt}. 
To evaluate the potential of these CL methods on our proposed CoSAM benchmark, we specifically modify them to make them applicable to our benchmark for continual adaptation as:}
\hazel{1) For LwF, when training on the new task, the knowledge from the old adapter trained on the previous task is distilled into the new adapter to alleviate the catastrophic forgetting; 2) For EWC, we utilize the importance-weighted discrepancies between the old and current adapters as a regularization to constrain the change of key parameters of previous tasks during training on new tasks; 3) For ER, we construct a memory bank to refresh the adapter trained on old tasks. For these classic CL methods, we adopt the adapter structure in HQ-SAM ~\cite{sam_hq} for SAM adaptation. 4) For L2P, we initialize a prompt pool with key-value token pairs and exploit the image features extracted by SAM image encoder to query the corresponding value token from the prompt pool for obtaining the task-specific information; 5) For Dual-Prompt, we use a general prompt to learn the task-sharing context, and expert prompts to learn the task-specific context; 6) For Coda-Prompt, it employs an attention mechanism for a soft selection of prompts. For Dual-Prompt and Coda-Prompt, we adopt similar modifications as for L2P to acquire the task-specific information for continual adaptation.}
\begin{figure*}[!t]
	\centering
	\includegraphics[width=0.9\textwidth]{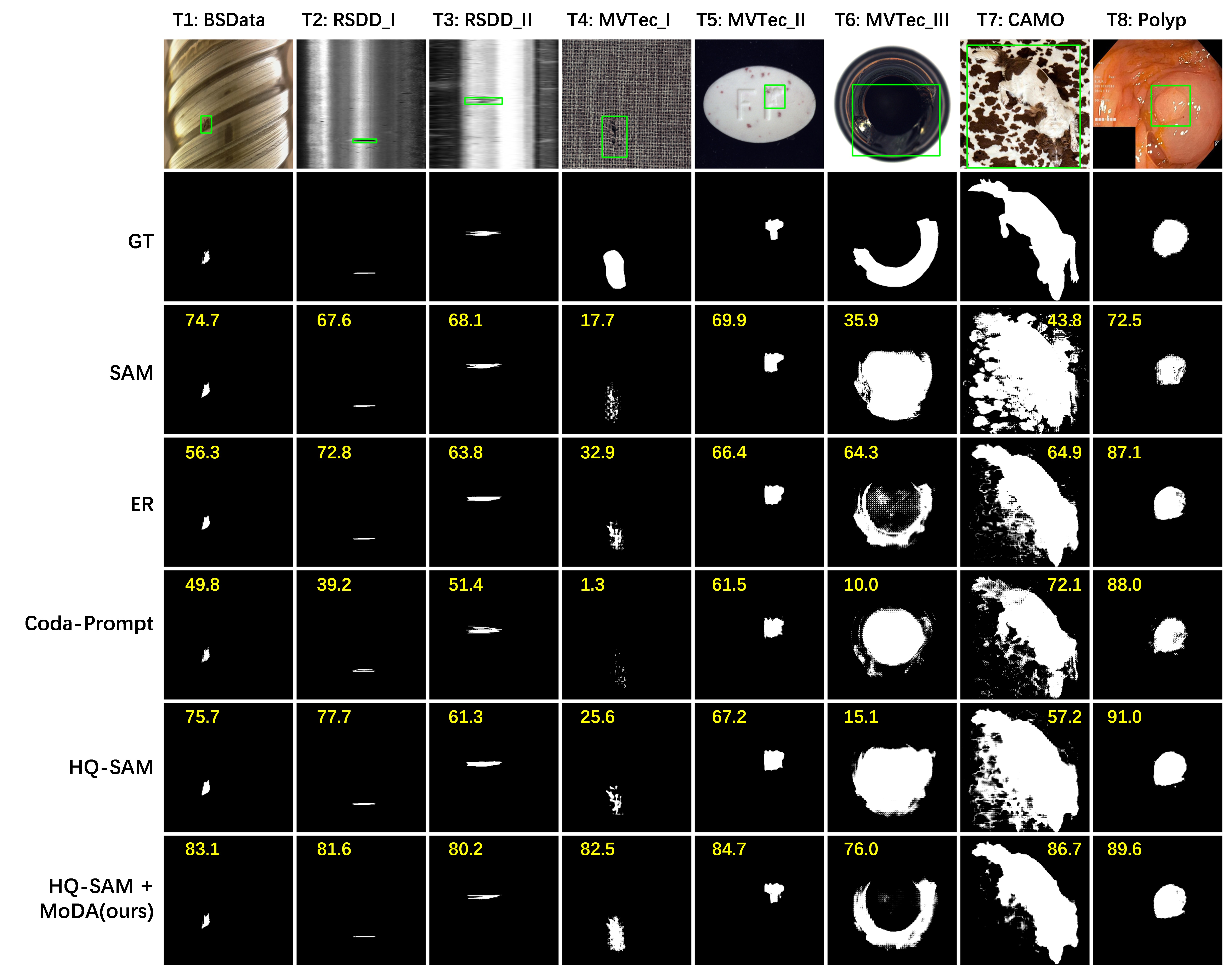}
	\caption{Qualitative comparison of competing methods with corresponding IoU values after continual training on all eight tasks. Please refer to supplementary materials for more results.}
        \label{fig:quality-compare-select}
        \vspace{-4mm}
\end{figure*}

\hazel{To validate the effectiveness of our proposed MoDA, several representative \jinglong{one-step adaptation} methods are adopted, including Decoder-Tuning~\cite{decoder-tuning}, SAM-Adapter~\cite{sam_adapter}, and HQ-SAM~\cite{sam_hq}. For continual adaptation, we execute these methods with a naive approach and our proposed MoDA, respectively. Here the naive manner represents that they are sequentially trained on every task without introducing any CL-related methodological design. For comprehensive evaluation, we will report the results of joint training, which is the upper bound, and the zero-shot performance of the original SAM without any adjustment.}

\subsection{Experimental Results}

\vspace{2mm}
\noindent\hazel{\textbf{Quantitative Evaluation.}
Table~\ref{tab:combined_performance_comparison} reports the quantitative results of different comparison methods on the proposed CoSAM benchmark. It is easily observed that 1) For adapter-based methods, including Decoder-Tuning, SAM-Adapter, and HQ-SAM, when adopting the naive sequential training strategy, they suffer from an extremely serious catastrophic forgetting with higher FF-IoU and FF-BIoU, and the average segmentation performance over all the 8 tasks Last-IoU is even inferior to the original SAM without any training procedure; 2) Compared to the naive training manner executed on HQ-SAM, the introduction of classic CL methods makes HQ-SAM consistently alleviate forgetting and helps achieve obvious performance improvement, outperforming the original SAM. This reflects the rationale for our modification of classic CL methods and tells us that properly introducing the CL technology into SAM can fully exploit the powerful capabilities of SAM to make it play a great role in streaming tasks; 3) For prompt-based CL methods, opposite to the performance trend for classification tasks, their performance is generally not as good as classic CL methods on the CoSAM benchmark for segmentation tasks. The underlying reason is analyzed in Sec.~\ref{sec:fail} above; 4) When incorporating the proposed MoDA into the adapter-based baselines, they consistently achieve significant performance improvement with stronger robustness and the catastrophic forgetting is largely alleviated. Especially, on the basis of HQ-SAM, the introduction of our MoDA makes it approach the upper bound.}


\begin{table*}[!t]
\centering
\caption{Comparison of Different Query Mechanism, with results from five seeds (mean and standard deviation).}
\vspace{-3mm}
\renewcommand{\arraystretch}{1}
\setlength{\tabcolsep}{0.4mm}{
\begin{tabular}{c|ccc|ccc|ccc}
\toprule
\multirow{2}{*}{Type of Pool} & \multicolumn{3}{c|}{Query}                           & \multicolumn{3}{c|}{IoU}                                    & \multicolumn{3}{c}{BIoU}                                   \\ \cline{2-10} 
                               & Mean Feature         & GFT           & GAT           & Last-IoU         & Avg-IoU          & FF-IoU          & Last-BIoU        & Avg-BIoU         & FF-BIoU         \\ \hline
\multirow{3}{*}{HQ-SAM}        & \checkmark           &               &               & 66.1 \scriptsize{$\pm$ 1.35} & 62.4 \scriptsize{$\pm$ 1.56} & 2.2 \scriptsize{$\pm$ 0.41} & 56.8 \scriptsize{$\pm$ 1.47} & 61.7 \scriptsize{$\pm$ 1.84} & 3.5 \scriptsize{$\pm$ 0.52} \\ 
                               &                       & \checkmark    &               & 72.8 \scriptsize{$\pm$ 0.68} & 70.7 \scriptsize{$\pm$ 0.42} & 1.2 \scriptsize{$\pm$ 0.32} & 63.1 \scriptsize{$\pm$ 0.59} & 69.6 \scriptsize{$\pm$ 0.93} & 1.7 \scriptsize{$\pm$ 0.28} \\ 
                               &                       & \checkmark    & \checkmark    & 75.2 \scriptsize{$\pm$ 0.42} & 74.9 \scriptsize{$\pm$ 0.61} & 0.28 \scriptsize{$\pm$ 0.09} & 65.3 \scriptsize{$\pm$ 0.37} & 71.6 \scriptsize{$\pm$ 0.50} & 0.14 \scriptsize{$\pm$ 0.07} \\ \hline
\multirow{3}{*}{L2P}           & \checkmark           &               &               & 63.1 \scriptsize{$\pm$ 1.15} & 56.5 \scriptsize{$\pm$ 1.49} & 2.0 \scriptsize{$\pm$ 0.43}   & 52.5 \scriptsize{$\pm$ 1.24} & 53.2 \scriptsize{$\pm$ 1.31} & 3.3 \scriptsize{$\pm$ 0.56} \\ 
                               &                       & \checkmark    &               & 66.9 \scriptsize{$\pm$ 0.58} & 64.6 \scriptsize{$\pm$ 0.53} & 1.1 \scriptsize{$\pm$ 0.20} & 56.1 \scriptsize{$\pm$ 1.35} & 59.8 \scriptsize{$\pm$ 1.47} & 1.5 \scriptsize{$\pm$ 0.24} \\ 
                               &                       & \checkmark    & \checkmark    & 68.3 \scriptsize{$\pm$ 0.35} & 66.7 \scriptsize{$\pm$ 0.49} & 0.31 \scriptsize{$\pm$ 0.13} & 56.9 \scriptsize{$\pm$ 0.52} & 61.4 \scriptsize{$\pm$ 0.48} & 0.26 \scriptsize{$\pm$ 0.10} \\ 
                               \bottomrule
\end{tabular}}
\vspace{-3mm}
\label{tab:ablat}
\end{table*}
\begin{table}[!t]
\centering
\caption{Effect of the length of Global Assistant Token (GAT) $N_a$ on HQ-SAM+MoDA.}
\vspace{-3mm}
\begin{tabular}{ccccccc}
\hline
\textbf{$N_a$} & \textbf{0} & \textbf{1} & \textbf{2} & \textbf{5} & \textbf{10} & \textbf{20} \\ \hline
\textbf{Last-IoU} & 72.8 & 73.6 & 74.1 & 74.8 & 75.2 & 75.2 \\
\textbf{Query Acc} & 69.6 & 81.7 & 88.2 & 94.6 & 98.1 & 98.5 \\ \hline
\end{tabular}
\label{tab:num-GAT}
\end{table}
\begin{table}[!t]
\centering
\caption{Effect of $|\mathcal{M}|$ on HQ-SAM+MoDA.}
\vspace{-2mm}

\begin{tabular}{ccccccc}
\hline
\textbf{$|\mathcal{M}|$} & \textbf{3} & \textbf{5} & \textbf{10} & \textbf{20}  \\ \hline
\textbf{Last-IoU} & 74.5 & 75.0 & 75.2 & 75.3  \\
\textbf{Query Acc} & 91.3 & 96.5 & 98.1 & 98.7 \\ \hline
\end{tabular}
\vspace{-4mm}
\label{tab:memory-bank-size}
\end{table}

\vspace{2mm}
\noindent\hazel{\textbf{Evaluation on Knowledge Preservation.}
After finishing the streaming learning on the CoSAM benchmark,
we further analyze the performance of the foundation model SAM on the natural image domain, \emph{i.e.}, the capability to retain general knowledge.}
\hazel{Table~\ref{tab:natural_domain} reports the quantitative results of different methods in the widely-adopted COCO dataset~\cite{coco}.
From it, we can easily observe that compared to other approaches, even after the sequential learning on a series of downstream tasks with quite different domain distributions, our proposed MoDA still helps achieve better segmentation performance on the natural image domain, approaching the original SAM. This fully demonstrates that our proposed MoDA has the stronger capability to preserve the general knowledge of SAM.
The main reason is that faced with such an obvious cross-domain testing scenario where the data distribution is quite different from that of CoSAM benchmark, the proposed GFT-based query mechanism can accurately help identify the take-related information (\emph{i.e.}, task ID) and effectively guides the model not to select any adapter from the pool trained on the previous 8 tasks with a higher probability. That is to say, during inference, our method has the capability to automatically switch between the original SAM and the adapter-based SAM depending on the incoming data domain. More verification about the proposed query mechanism is provided below.}

\vspace{2mm}
\noindent\hazel{\textbf{Visual Evaluation.}
Fig.~\ref{fig:quality-compare-select} shows the visual results predicted by different comparison methods after completing the streaming training on all the 8 tasks. To facilitate comparison, we provide the IoU score. As seen, although these comparison methods perform quite competitively on the last polyp segmentation task, they are inferior to the original SAM on most of previous tasks, showing an obvious forgetting phenomenon. 
However, when incorporating our proposed MoDA, this adapter-based approach HQ-SAM achieves a substantial improvement and obtains higher segmentation accuracy on all the 8 tasks. Besides, it is worth mentioning that for particularly difficult samples, such as targets that are not easy to identify as displayed in Tasks 4 and 6, our proposed method can always achieve very competitive results, showing strong application potential.}

\subsection{More Experimental Analysis}\label{sec:abla}

\vspace{2mm}
\noindent\hazel{\textbf{Ablation Study on GFT and GAT.} 
To further substantiate the rationality of our proposed query mechanism, we take L2P and HQ-SAM as the prompt pool and the adapter pool, respectively, and execute a series of ablation studies as listed in Table~\ref{tab:ablat}. Here, ``mean feature'' represents that the query is obtained by averaging the feature extracted by SAM encoder at the spatial dimension. Compared to the mean feature-based query mechanism, only introducing a simple global feature token (GFT) brings obvious performance gains for both prompt-based L2P and adapter-based HQ-SAM in the sequential learning scenario. With the help of global assistant tokens (GAT), the model's plasticity and ability to alleviate forgetting can be further improved, showing higher Last-IoU and lower FF-IoU scores. }

\vspace{2mm}
\noindent\hazel{\textbf{Effect of The Length of GAT $N_{a}$.} Table~\ref{tab:num-GAT} reports the influence of the length of GAT $N_{a}$ on the performance of HQ-SAM+MoDA. As seen, with the increase of $N_{a}$, the classification accuracy of querying the adapter pool for identifying different task domains is higher, which then boosts better segmentation performance. Especially, we can find that when $N_{a}$ varies from 0 to 1, the performance can be improved significantly, which finely demonstrates that GAT does play an important role in improving model plasticity. In experiment comparison, we set $N_{a}$ to 10.}

\vspace{2mm}
\noindent\hazel{\textbf{Effect of The Size of Memory Bank $|\mathcal{M}|$.} Table~\ref{tab:memory-bank-size} lists the results of HQ-SAM+MoDA under different $|\mathcal{M}|$ for storing the samples from different tasks. With the increase of the buffer size, the performance of previous tasks can be better maintained and the forgetting rate presents a downtrend. We finally set $|\mathcal{M}|$ to 10. Traditional replayed-based CL methods typically require 50-200 exemplars per task~\cite{der, er}.}


\section{Conclusion}

\hazel{In this paper, for the practical continual learning scenario, we introduced a novel Continual SAM adaptation (CoSAM) benchmark with 8 different task domains to assess the existing SAM adaptation methods. We comprehensively analyzed the limitations of SAM one-step adaptation methods for the segmentation task, and proposed a simple-yet-effective continual SAM adaptation algorithm, called Mixture of Domain Adapters (MoDA). Extensive experiments substantiated that our proposed MoDA has favorable versatility and robustness, and it consistently promoted the existing methods to achieve better segmentation performance with lower forgetting risk.
To our knowledge, we are the first to explore the integration of continual learning into the field of SAM adaptation. Such a promising exploration attempt confirms the feasibility of extending SAM to the dynamic streaming learning scenarios for segmentation and lays the foundation for future research in this field.}
{
    \small
    \bibliographystyle{ieeenat_fullname}
    \bibliography{main}
}


\clearpage
\setcounter{page}{1}
\maketitlesupplementary

\appendix
\begin{abstract}
In this supplementary material, we provide additional visual examples from our proposed CoSAM Benchmark in Section \ref{CoSAM Datasets}. Detailed implementation pipelines for adapting each comparison method to the proposed CoSAM benchmark are outlined in Section \ref{app:algorithms}. Section \ref{sec:across-train} provides more quantitative comparisons to evaluate the performance of different methods on seen tasks during the sequential training procedure. Section \ref{reverse-order} evaluates the task-order robustness of different methods on the CoSAM Benchmark by reversing the task order. Finally, in Section \ref{visual-compare}, we provide more visual results as well as IoU on diverse samples after training on the CoSAM Benchmark. Code is available at \href{https://github.com/yangjl1215/CoSAM}{https://github.com/yangjl1215/CoSAM}.
\end{abstract}

\section{Datasets in CoSAM Benchmark}
\label{CoSAM Datasets}
Our proposed CoSAM Benchmark is composed of different datasets from diverse domains, moving beyond the homogeneous domains of traditional continual learning tasks. This diversity allows us to comprehensively evaluate the effectiveness of continual adaptation methods in realistic scenarios characterized by dynamic and varied distributions. The CoSAM benchmark consists of five datasets which is divided into 8 tasks and 
there are clear differences between these 8 tasks.
\begin{itemize}[leftmargin=3.5mm, itemsep=0mm, topsep=0.5 mm]
    \item[$\circ$]BSData \cite{BSData}: Images of surface defects on industrial machine tool components for automated inspection.
    \item[$\circ$]RSDDs \cite{RSDD}: Railway track surface defects are divided into two tasks, representing distinct domains of high-speed and heavy-duty transport tracks, as shown in Figs.~\ref{fig-rsdd} and \ref{fig-rsdd2}.  
    \item[$\circ$]MVTec Anomaly Segmentation \cite{mvtec}: Containing 15 objects with over 70 defect types, which is reorganized into three tasks to mitigate data imbalance, as displayed in Figs.~\ref{fig-mvtec}, \ref{fig-mvtec2}, and \ref{fig-mvtec3}.    
    \item[$\circ$]CAMO \cite{camo}: Camouflaged object segmentation for applications in medicine, agriculture, and art.
    \item[$\circ$]Polyp Segmentation \cite{polyp}: Medical imaging dataset for detecting polyps, critical in colorectal cancer prevention.
  \end{itemize}

\begin{figure}[H]
	\centering
	\includegraphics[width=0.45\textwidth]{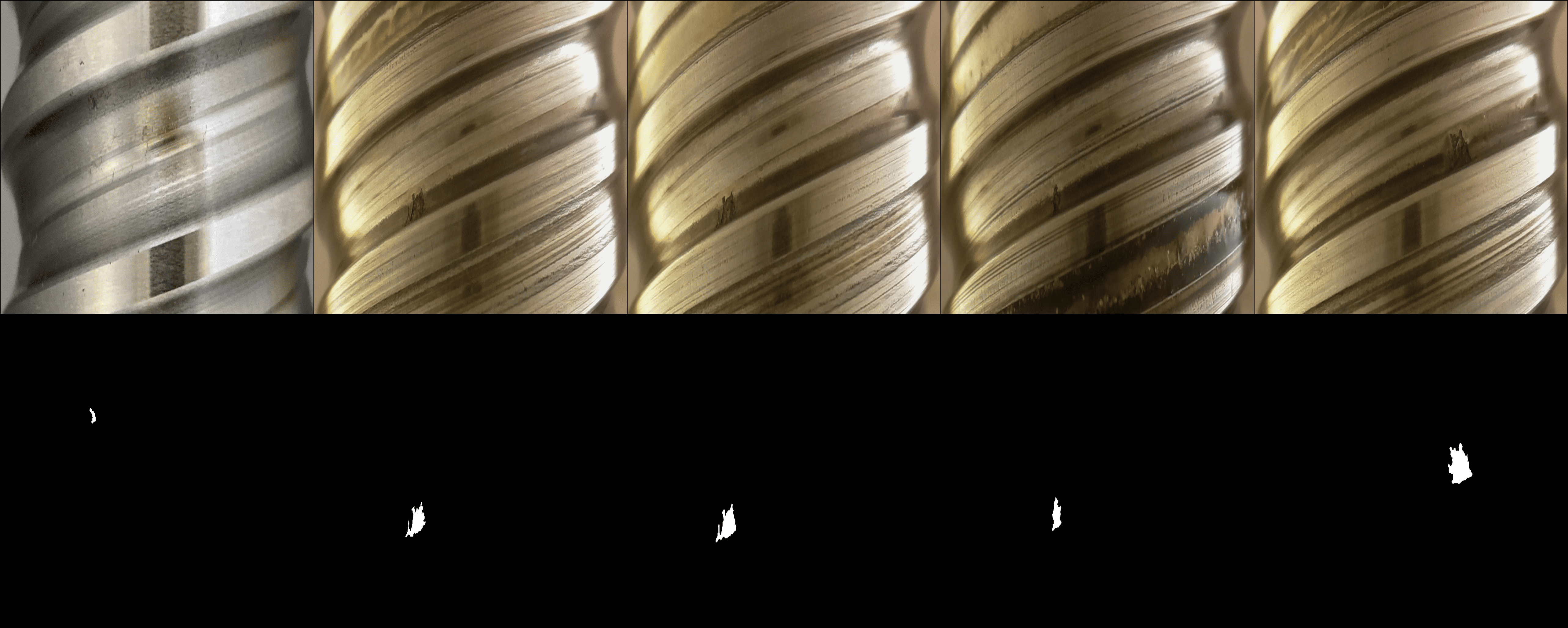}
	\caption{Task1: BSData}
        \label{fig-bsdata}
\end{figure}

\begin{figure}[H]
	\centering
	\includegraphics[width=0.45\textwidth]{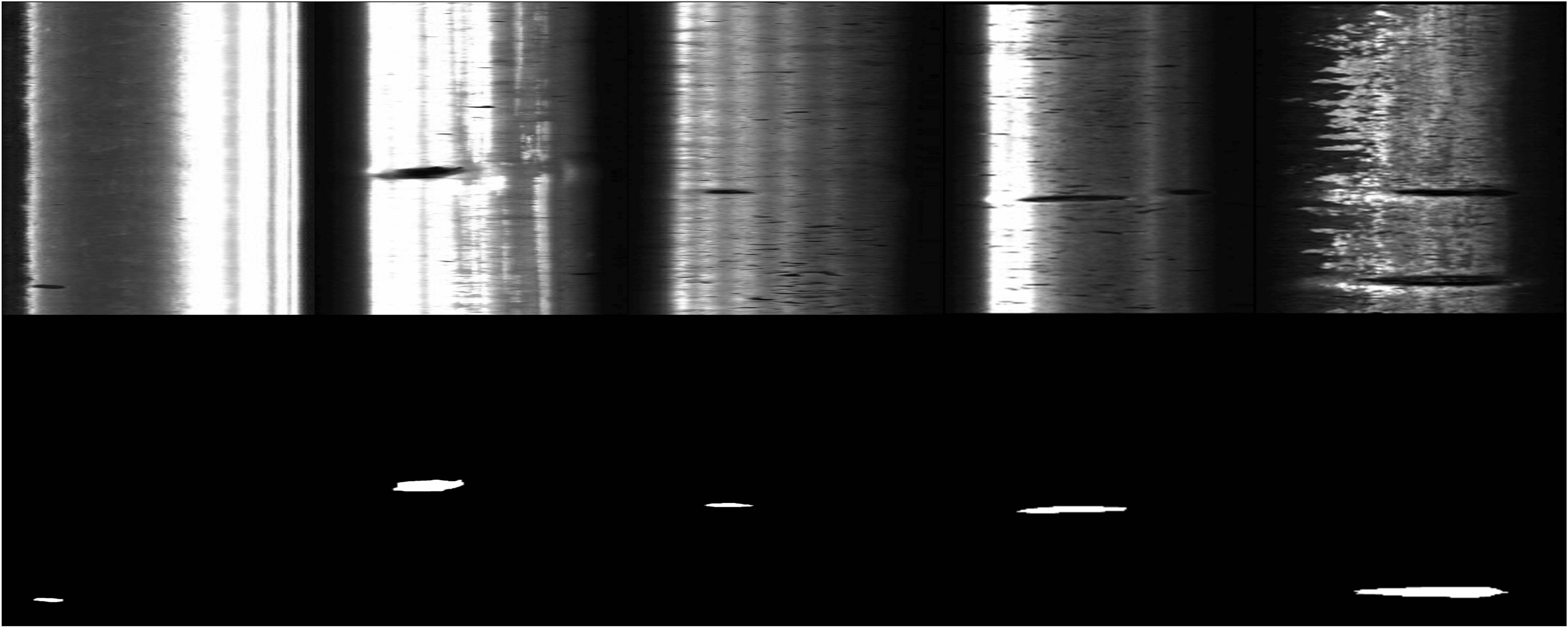}
	\caption{Task2: RSDD\_I}
        \label{fig-rsdd}
\end{figure}

\begin{figure}[H]
	\centering
	\includegraphics[width=0.45\textwidth]{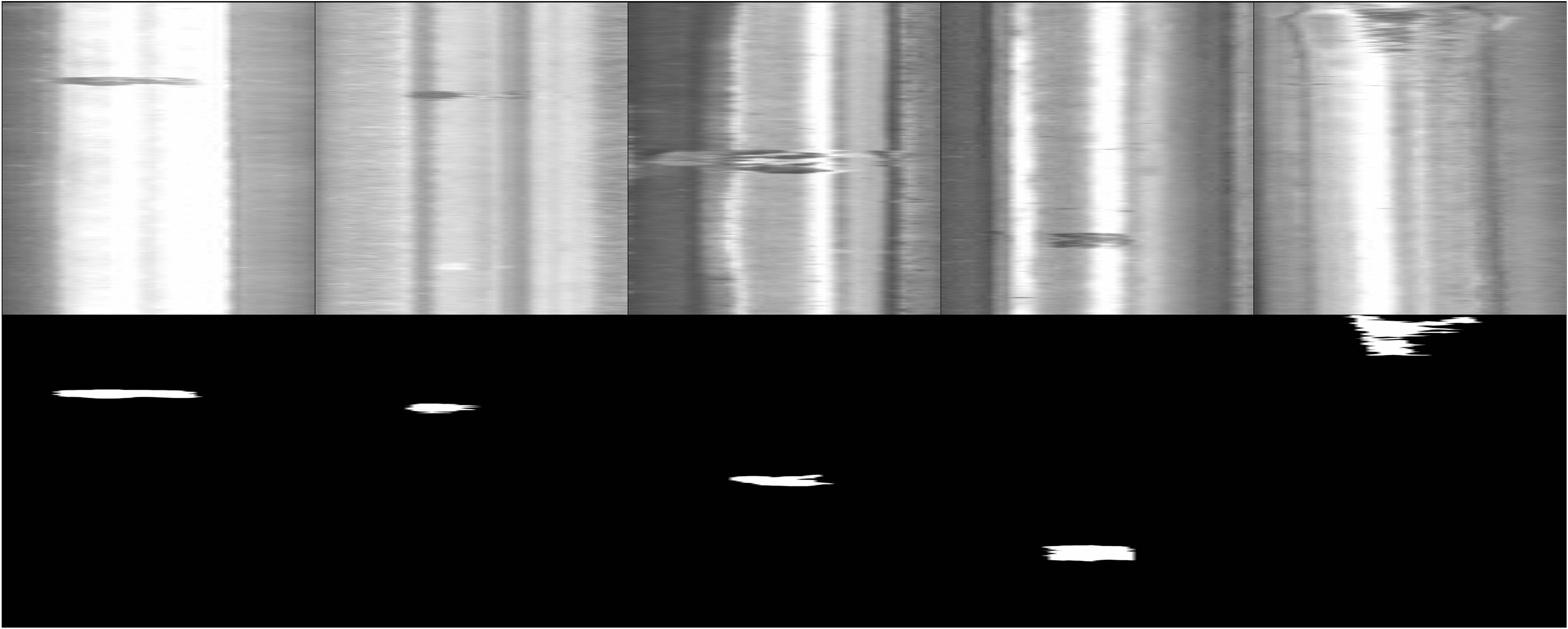}
	\caption{Task3: RSDD\_{II}}
        \label{fig-rsdd2}
\end{figure}

\begin{figure}[H]
	\centering
	\includegraphics[width=0.45\textwidth]{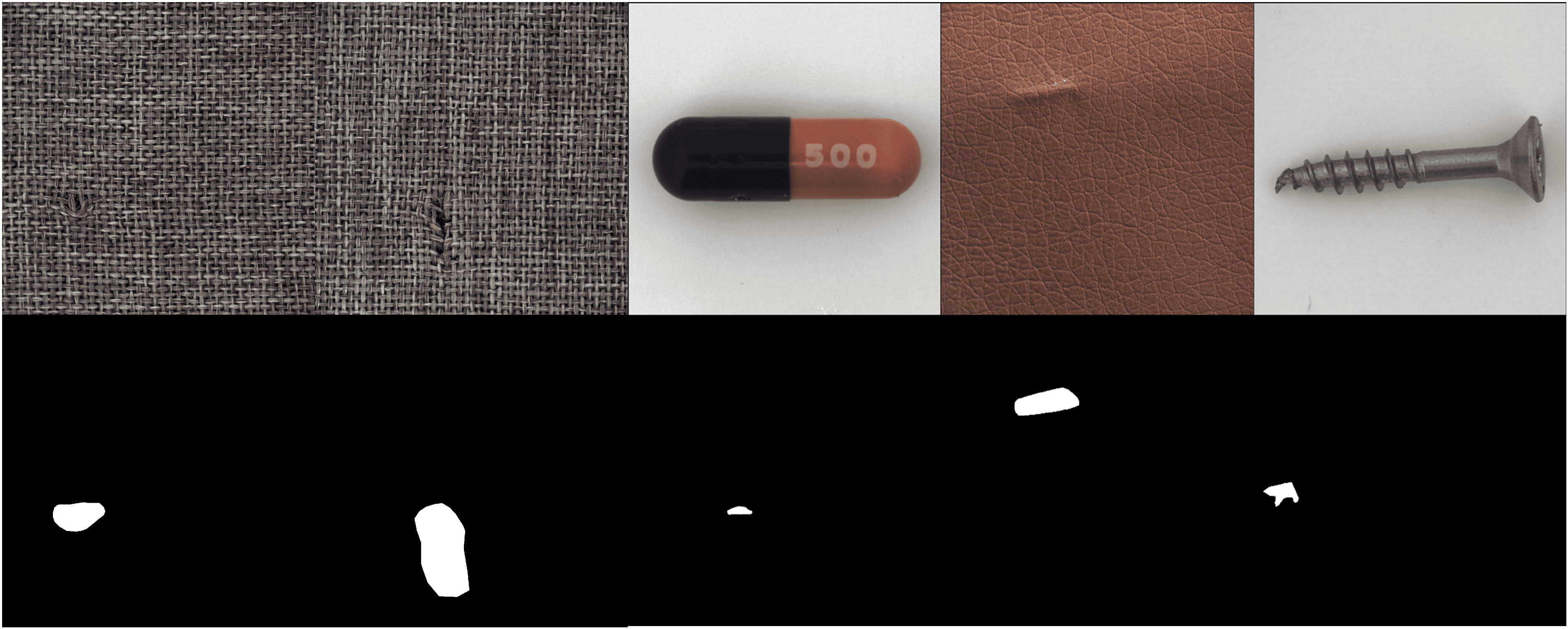}
	\caption{Task4: MVTec\_I}
        \label{fig-mvtec}
\end{figure}

\begin{figure}[H]
	\centering
	\includegraphics[width=0.45\textwidth]{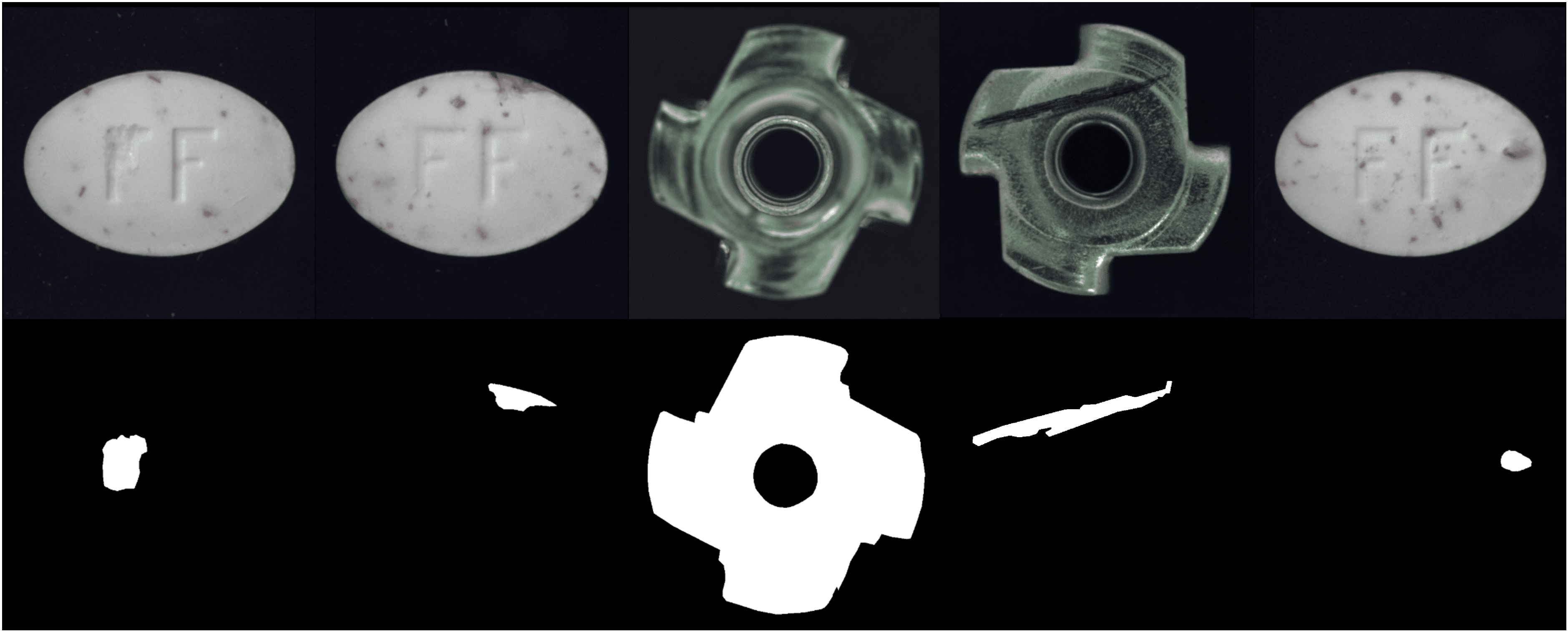}
	\caption{Task5: MVTec\_{II}}
        \label{fig-mvtec2}
\end{figure}

\begin{figure}[H]
	\centering
	\includegraphics[width=0.45\textwidth]{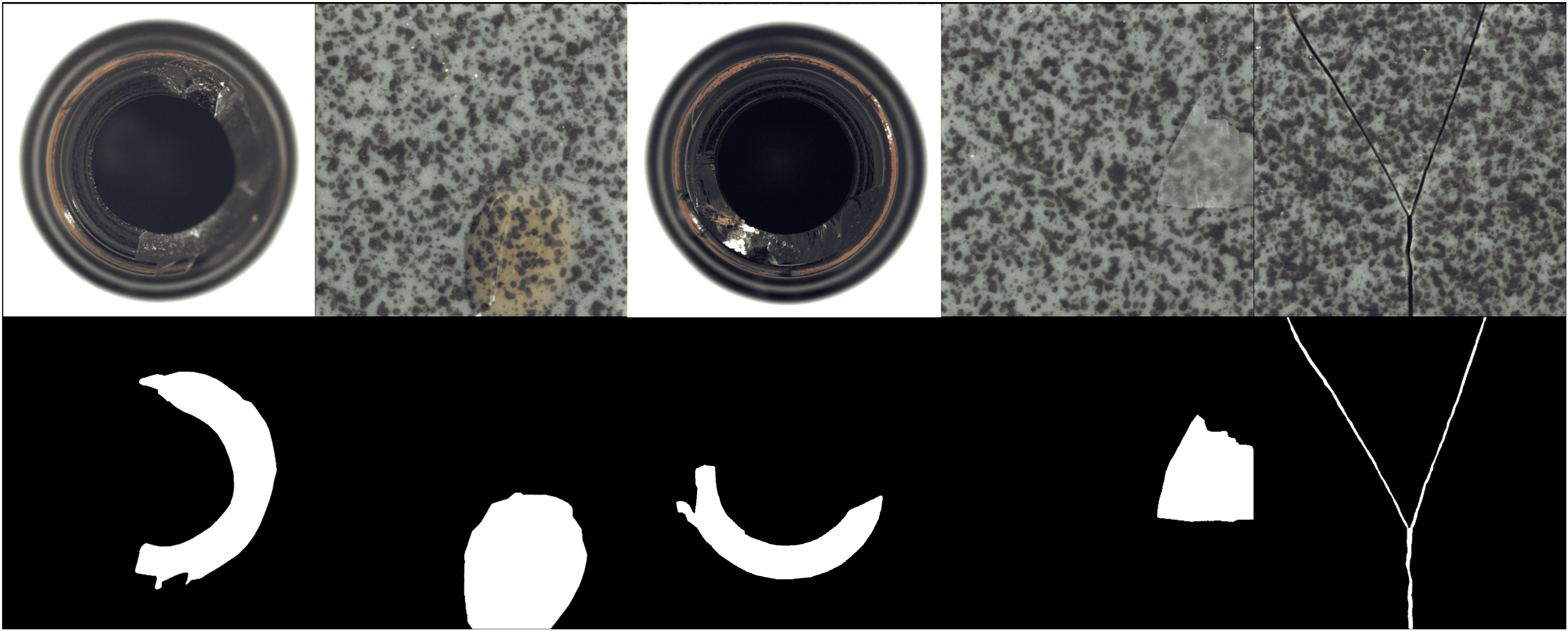}
	\caption{Task6: MVTec\_{III}}
        \label{fig-mvtec3}
\end{figure}

\begin{figure}[H]
	\centering
	\includegraphics[width=0.45\textwidth]{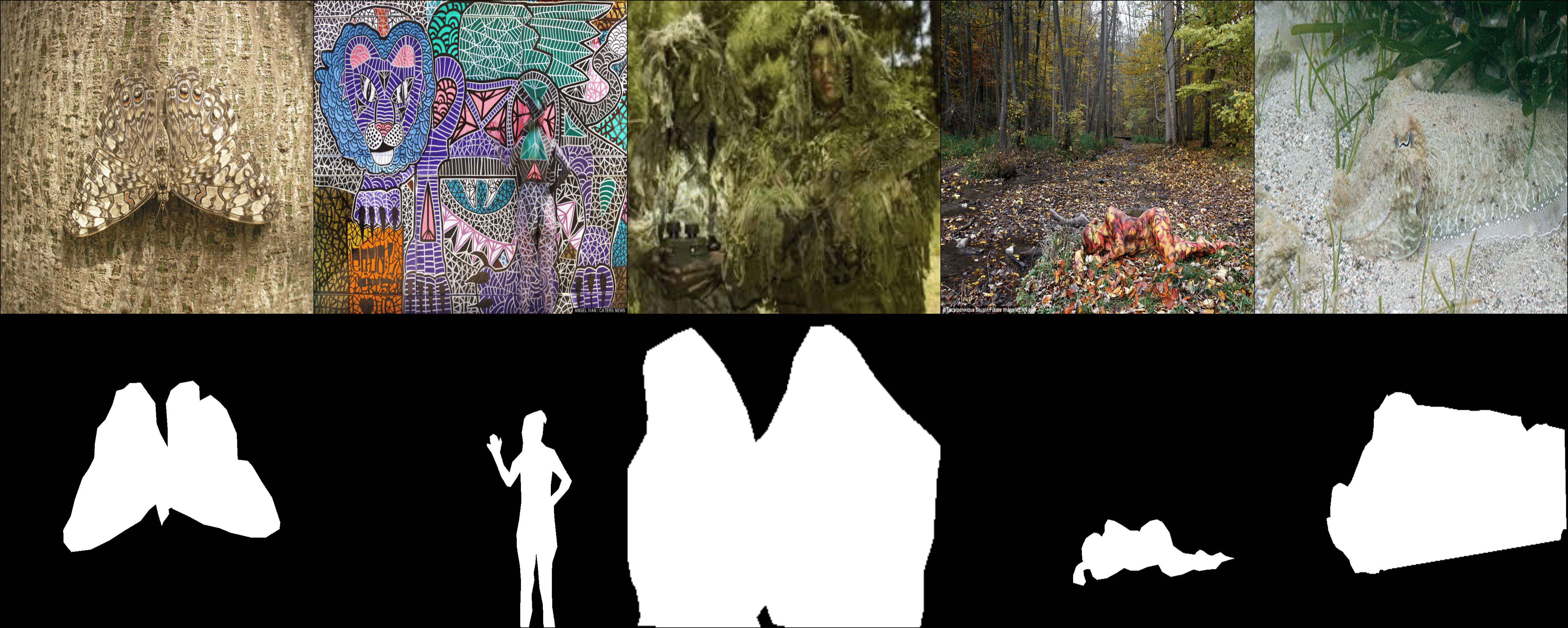}
	\caption{Task7: CAMO}
        \label{fig-camo}
\end{figure}

\begin{figure}[H]
	\centering
	\includegraphics[width=0.45\textwidth]{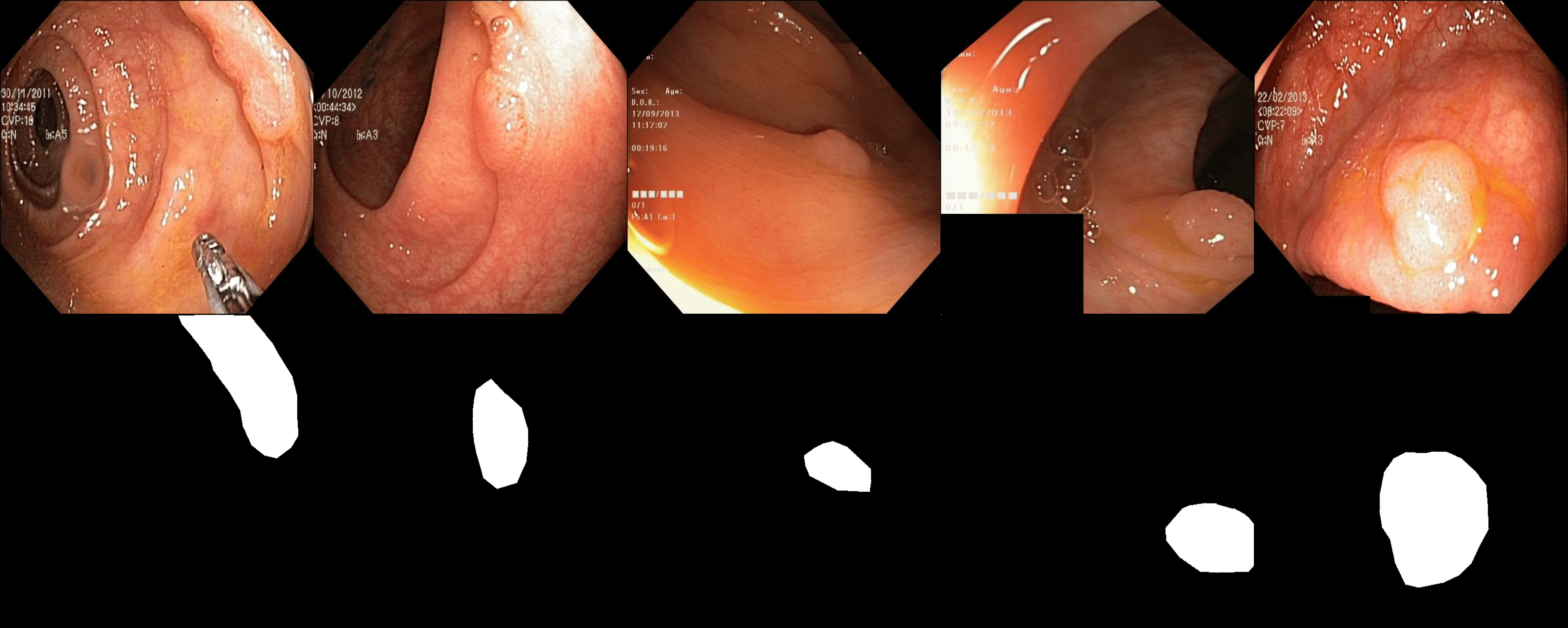}
	\caption{Task8: Polyp}
        \label{fig-polyp}
\end{figure}

\section{Pipeline of Algorithms}
\label{app:algorithms}

\paragraph{Continual Adaptation with LwF} Learning without Forgetting (LwF)\cite{lwf} employs knowledge distillation to ensure that the new model does not deviate substantially from its predecessor. This approach retains a copy of the previous model; during training on the current task, the input is processed by the previous model to extract representative features containing the knowledge from earlier tasks. This extracted knowledge is then distilled into the current model to preserve learned capabilities. In our continual adaptation setting, we maintain the adapter from the old task. When training a new adapter, the knowledge from the old adapter is distilled into the new adapter, effectively integrating past learning into the ongoing adaptation process. This method is illustrated in Algorithm \ref{alg:lwf}.

\begin{algorithm}[t]
    \caption{Continual Adaptation with LwF}\label{alg:lwf}
    \begin{algorithmic}[1]
        \REQUIRE{Pre-trained SAM $\theta$, Tasks $\mathcal{D}=\{D_1,\dots,D_N\}$}
        \ENSURE{Adapter $\phi$}
        \STATE Initialize Adapter $\phi_1$.  
        \FOR{$x, y \in D_1$} 
            \STATE{Update $\phi_1$ with $\mathcal{L}_{\text{mask}}((f_{\theta} (x, \phi_1) , y)$.}
        \ENDFOR
        \FOR {$D_t \in \{D_2, \dots, D_N \}$}
            \STATE Initialize Adapter $\phi_t$.  
            \FOR{$x, y \in D_t$}
                \STATE{Update Adapter $\phi_t$ with $\mathcal{L}_{\text{mask}}(f_{\theta} (x, \phi_t) , y) + \gamma \mathcal{L}_{\text{mask}}(f_{\theta} (x, \phi_t) , f_{\theta} (x, \phi_{t-1}))$.}
            \ENDFOR
        \ENDFOR
        \RETURN $\phi_N$
    \end{algorithmic}
\end{algorithm}

\paragraph{Continual Adaptation with EWC} Elastic Weight Consolidation (EWC) \cite{ewc}, conserves both the previous model and the significance of each parameter, as assessed by the Fisher information matrix. This method uses the importance-weighted discrepancies between the old and current models as a constraint. This ensures that critical parameters from earlier tasks undergo minimal changes during updates for new tasks, thereby preserving the essential knowledge needed to address previous challenges. We use EWC to update the weights of Adapter in the CoSAM Benchmark. Detailed procedure is described in Algorithm \ref{alg:ewc}.

\begin{algorithm}[t]
    \caption{Continual Adaptation with EWC}\label{alg:ewc}
    \begin{algorithmic}[1]
        \REQUIRE{Pre-trained SAM $\theta$, Tasks $\mathcal{D}=\{D_1,\dots,D_N\}$}
        \ENSURE{Adapter $\phi$}
        \STATE Initialize Adapter $\phi_1$.  
        \FOR{$x, y \in D_1$} 
            \STATE{Update $\phi_1$ with $\mathcal{L}_{\text{mask}}((f_{\theta} (x, \phi_1) , y)$.}
        \ENDFOR
        \STATE Calculate Fisher Information Matrix $F_1$ for $\phi_1$.
        \FOR {$D_t \in \{D_2, \dots, D_N \}$}
            \STATE Initialize $\phi_t$.  
            \FOR{$x, y \in D_t$}
                \STATE{Update $\phi_t$ with $\mathcal{L}_{\text{mask}}(f_{\theta} (x, \phi_t) , y) + \beta \sum_i F_{t-1,i} \cdot (\phi_{t, i} - \phi_{t-1, i})$.}
            \ENDFOR
            \STATE Calculate $F_t$ for $\phi_t$.
        \ENDFOR
        \RETURN $\phi_N$
    \end{algorithmic}
\end{algorithm}

\paragraph{Continual Adaptation with ER} Experience Replay (ER) \cite{er} is a simple yet effective method to prevent catastrophic forgetting. It eases the constraint on accessing old data by permitting the use of a retained dataset. In the CoSAM Benchmark, We construct a small memory bank to train the Adapter to enhance its efficacy on older tasks. The specific steps are described in Algorithm \ref{alg:er}.

\begin{algorithm}
    \caption{Continual Adaptation with ER}\label{alg:er}
    \begin{algorithmic}[1]
        \REQUIRE{Pre-trained SAM $\theta$, Tasks $\mathcal{D}=\{D_1,\dots,D_N\}$}
        \ENSURE{Adapter $\phi$, Memory Bank $\mathcal{M}$}
        \STATE Initialize Adapter $\phi$.
        \FOR{$x, y \in D_1$}
            \STATE{Update $\phi$ with $\mathcal{L}_{\text{mask}}((f_{\theta} (x, \phi_1) , y)$.}
        \ENDFOR
        \STATE Select exemplars into $\mathcal{M}$.
        \FOR {$D_t \in \{D_2, \dots, D_N \}$}
            \FOR{$x, y \in D_t \cup \mathcal{M}$}
                \STATE{Update $\phi$ with $\mathcal{L}_{\text{mask}}((f_{\theta} (x, \phi_1) , y)$.}
            \ENDFOR
            \STATE Select exemplars into $\mathcal{M}$.
        \ENDFOR
        \RETURN $\phi$, $\mathcal{M}$
    \end{algorithmic}
\end{algorithm}

\paragraph{Continual Adaptation with L2P} Learning to Prompt (L2P) \cite{l2p} starts the paradigm of using learnable tokens to tune a pre-trained model in Continual Learning. In our CoSAM Benchmark, we init a Prompt Pool $\mathcal{P}$ with Key-Value token pairs $\{K:V\}$. The SAM image encoder $\theta'$ extracts the image patch token $P_{img}$ and calculates the spatial mean $z$ as the image feature, which is then used to query the nearest key token. Then the corresponding value token $v$ is concatenated with $P_{img}$ for the further process by SAM's decoder. The detailed procedure is shown in Algorithm \ref{alg:l2p}.

\begin{algorithm}[t]
    \caption{Continual Adaptation with L2P}\label{alg:l2p}
    \begin{algorithmic}[1]
        \REQUIRE{Pre-trained SAM Encoder $\theta'$ and Decoder $\psi$, Tasks $\mathcal{D}=\{D_1,\dots,D_N\}$}
        \ENSURE{$\mathcal{P}$=\{Key Tokens $K$: Value Tokens $V$\}}
        \STATE Initialize $\mathcal{P} = \{K: V\}$.
        \FOR {$D_t \in \{D_1, \dots, D_N \}$}
            \FOR{$x, y \in D_t$}
                \STATE{$P_{img} = f_{\theta'}(x)$}
                \STATE $z = mean(P_{img})$
                \STATE Matched prompt $ id = \argmin_{i} MSE(K[i], z)$.
                \STATE Value token $v = V[id]$.
                \STATE Concatenate $v$ to $P_{img}$ and update $v$ with $\mathcal{L}_{\text{mask}}(f_{\psi} (\left[v;P_{img}\right]) , y)$.
                \STATE Update matched key by minimizing $MSE(K[id],z)$.
            \ENDFOR
        \ENDFOR
        \RETURN Prompt Pool $\mathcal{P}$
    \end{algorithmic}
\end{algorithm}

\paragraph{Continual Adaptation with Dual-Prompt} Dual-Prompt \cite{dualprompt} is an extension of L2P. For SAM Continual Adaptation, we initiate General Prompt Token $g$ and Expert Prompt Pool $\mathcal{P}$ with Key-Value token pairs $\{K:V\}$. The SAM image encoder extracts the image feature $z$, which is used to query the Expert Prompt Pool, then the matched value token $v$ is concatenated with $P_{img}$ for the further processing by SAM's decoder. Specifically, $g$ is concatenated to the shallow feature sequence, whereas the expert prompt is concatenated to the deep feature sequence. This design can help the general prompt to be updated for task-sharing context, while the expert prompt to be updated for task-specific context. The detailed procedure is shown in Algorithm \ref{alg:dualprompt}.

\begin{algorithm}[t]
    \caption{Continual Adaptation with Dual-Prompt}\label{alg:dualprompt}
    \begin{algorithmic}[1]
        \REQUIRE{Pre-trained SAM Encoder $\theta'$ and Decoder $\psi$, Tasks $\mathcal{D}=\{D_1,\dots,D_N\}$}
        \ENSURE{General Prompt Token $g$, Expert Prompt Pool $\mathcal{P}$=\{Key Tokens $K$: Value Tokens $V$\}}
        \STATE Initialize $g$, $\mathcal{P} = \{K: V\}$.
        \FOR {$D_t \in \{D_1, \dots, D_N \}$}
            \FOR{$x, y \in D_t$}
                \STATE{$P_{img} = f_{\theta'}(x)$}
                \STATE $z = mean(P_{img})$
                \STATE Matched prompt $ id = \argmin_{i} MSE(K[i], z)$.
                \STATE Value token $v = V[id]$.
                \STATE Concatenate $g$ to the shallow feature sequence, $v$ to the deep feature sequence, then update $g,v$ with $\mathcal{L}_{\text{mask}}(f_{\psi} (\left[g,v;P_{img}\right]) , y)$.
                \STATE Update matched key by minimizing $MSE(K[id],z)$.
            \ENDFOR
        \ENDFOR
        \RETURN General Prompt $g$, Expert Prompt Pool $\mathcal{P}$
    \end{algorithmic}
\end{algorithm}

\paragraph{Continual Adaptation with Coda-Prompt \cite{codaprompt}} In addition to segmentation loss, methods like L2P need an additional key matching loss to align the key tokens with the image features, which introduces additional complexit. In contrast, Coda-Prompt addresses this issue by employing an attention mechanism to perform a soft selection of value tokens. Specifically, for SAM Continual Adaptation, the image feature $z$ extracted by the SAM encoder is multiplied with all key tokens in the prompt pool. The resulting values are then processed through a softmax layer to produce weights for a weighted sum of the value tokens. The detailed procedure is outlined in Algorithm \ref{alg:codaprompt}.

\begin{algorithm}[t]
    \caption{Continual Adaptation with Coda-Prompt}\label{alg:codaprompt}
    \begin{algorithmic}[1]
        \REQUIRE{Pre-trained SAM Encoder $\theta'$ and Decoder $\psi$, Tasks $\mathcal{D}=\{D_1,\dots,D_N\}$}
        \ENSURE{Prompt Pool $\mathcal{P}$=\{Key Tokens $K$: Value Tokens $V$\}}
        \STATE Initialize $\mathcal{P} = \{K: V\}$.
        \FOR {$D_t \in \{D_1, \dots, D_N \}$}
            \FOR{$x, y \in D_t$}
                \STATE{$P_{img} = f_{\theta'}(x)$}
                \STATE $z = mean(P_{img})$
                \STATE $w = softmax(z \times K)$
                \STATE Concatenate weighted $V$ and update $K,V$ with $\mathcal{L}_{\text{mask}}(f_{\psi} (\left[w \odot V;P_{img}\right]) , y)$.
            \ENDFOR
        \ENDFOR
        \RETURN Prompt Pool $\mathcal{P}$
    \end{algorithmic}
\end{algorithm}

\section{More Performance Evaluation Across the Sequential Training}
\label{sec:across-train}
\begin{figure}[t]
	\centering
	\includegraphics[width=0.5\textwidth]{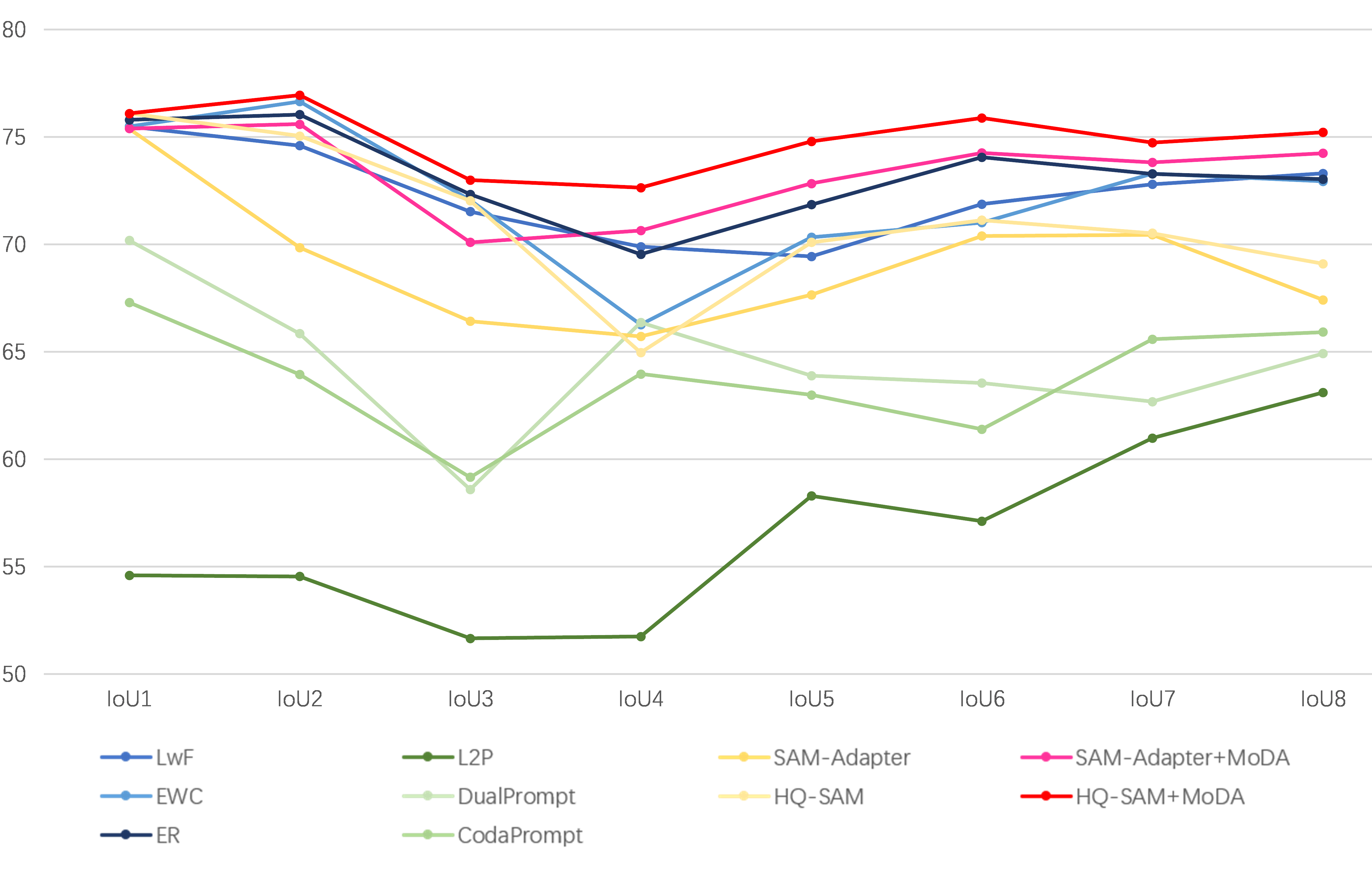}
	\caption{Quantitative comparison across the training procedure. $IoU_t$ represents the average IoU performance across all test sets of the first $t$ tasks after training on the $t$-th task.}
        \label{fig:avgiou}
\end{figure}

\begin{figure}[t]
	\centering
	\includegraphics[width=0.5\textwidth]{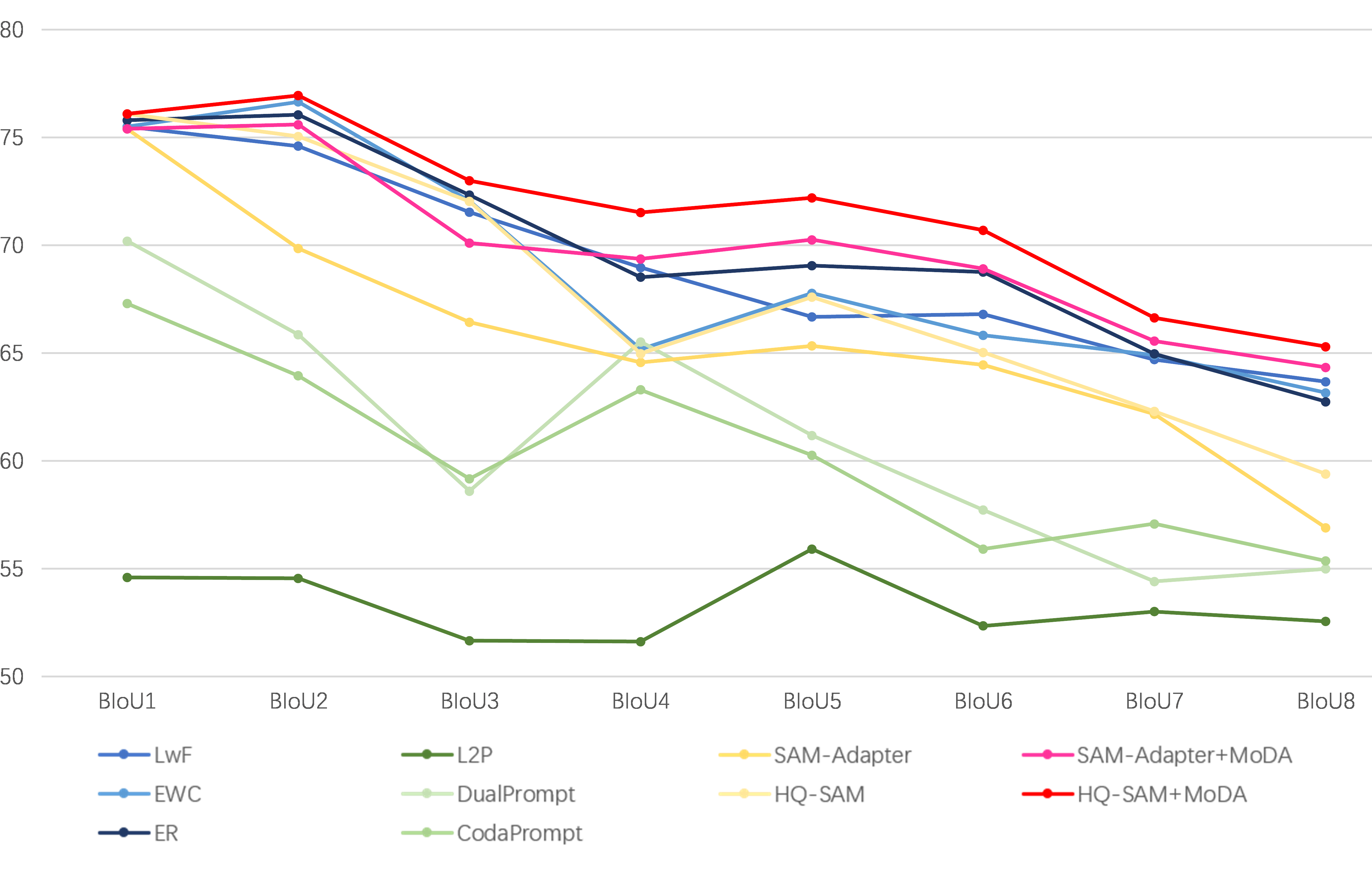}
	\caption{Quantitative comparison across the training procedure. $BIoU_t$ represents the average BIoU performance across all test sets of the first $t$ tasks after training on the $t$-th task.}
        \label{fig:avgbiou}
\end{figure}

We provide more quantitative results of different methods 
across the entire training stream. Fig. \ref{fig:avgiou} and Fig. \ref{fig:avgbiou} report the $IoU_{t}$ and $BIoU_t$, which represent the average IoU and BIoU on all seen tasks after training on task $t$, respectively. As observed, for prompt-based CL methods, such as L2P, DualPrompt, and CodaPrompt, they struggle significantly throughout the process due to their inability to match task-specific prompt tokens. For one-step adaptation methods, such as HQ-SAM and SAM-Adapter, they suffer from a certain degree of forgetting effect. For classical continual learning methods, such as LwF, ER, and EWC, benefiting from their effective forgetting mitigation strategies, they achieve promising results. Compared to these methods, the introduction of our proposed MoDA significantly alleviates the forgetting effect and then helps achieve superior performance.

\section{Evaluation on Task-Order Robustness}
\label{reverse-order}

To further evaluate the effectiveness of our proposed MoDA, 
we add an experimental setting where the task order is reversed. In this scene, the first task BSDdata is converted to the eighth task. Table \ref{tab:reverse_order} reports the IoU on every task after finishing the sequential training on the 8 tasks.
From it, we can easily find that 1) when the task order is reversed, for LwF and HQ-SAM, although the performance on latest tasks, \emph{e.g.}, BSDdata and RSDD\_I, presents certain improvement, the performance on previous tasks consistently shows an obvious downtrend. For example, for HQ-SAM, the IoU on Polyp varies from 82.1\% to 22.7\%. These results show that LwF and HQ-SAM suffer from forgetting; 2) For L2P, it exhibits certain performance fluctuations. For many tasks, such as MVTec\_II, MVTec\_III, and Polyp, the performance is actually worse when these tasks appear later in the sequence. This inconsistency stems from the inability to effectively match corresponding task prompts, leading to chaotic prompt selection during both training and inference, regardless of the task order; 3) With the introduction of our proposed MoDA, 
no matter what the order of tasks is, HQ-SAM presents similar performance on every task. This is mainly attributed to the fact that our proposed query mechanism is not affected by the task order and it can always provide the accurate task-specific information for correctly matching the corresponding adapter.





\begin{table*}[]
\caption{Performance evaluation on task-order robustness. IoU score is reported for each task.}
\label{tab:reverse_order}
\resizebox{\textwidth}{!}{
\begin{tabular}{@{}lllllllll@{}}
\toprule
\textbf{Method}                     & \textbf{T1/8: BSData}  & \textbf{T2/7: RSDD\_I} & \textbf{T3/6: RSDD\_II} & \textbf{T4/5: MVTec\_I} & \textbf{T5/4: MVTec\_II} & \textbf{T6/3: MVTec\_III} & \textbf{T7/2: CAMO}    & \textbf{T8/T1: Polyp}  \\ \midrule
LwF                  & 71.6          & 71.2          & 63.6           & 68    & 83.8   & 73.9    & 73.6 & 80.8 \\
LwF(reverse)         & 74.2 & 73.2 & 66.4  & 63.2           & 79.2            & 70.4             & 63.7          & 68.8          \\
\cline{1-9}
L2P                  & 67.6          & 69.3          & 60.1  & 49.1           & 76.3            & 53.6             & 61.9 & 67            \\
L2P(reverse)         & 72.8 & 71.3 & 59.3           & 53.2  & 78.3   & 60      & 57.9          & 68.8 \\
\cline{1-9}
HQ-SAM               & 70.3          & 65.4          & 57.9           & 60.9  & 80.9   & 69.9    & 65.7 & 82.1 \\
HQ-SAM(reverse)      & 76.9 & 76.5 & 68.7  & 58.8           & 71.9            & 61.2             & 49.5          & 22.7          \\
\cline{1-9}
HQ-SAM+MoDA          & 76.1          & 77.8          & 65.2           & 71.6           & 83.3            & 81.3             & 67.7          & 78.8          \\
HQ-SAM+MoDA(reverse) & 75.8          & 77.1          & 65.9           & 71.7           & 84.3            & 81.2             & 68.4          & 79            \\ \bottomrule
\end{tabular}
}
\end{table*}

\section{More Visual Results}
\label{visual-compare}
\begin{figure*}[t]
	\centering
	\includegraphics[width=0.8\textwidth]{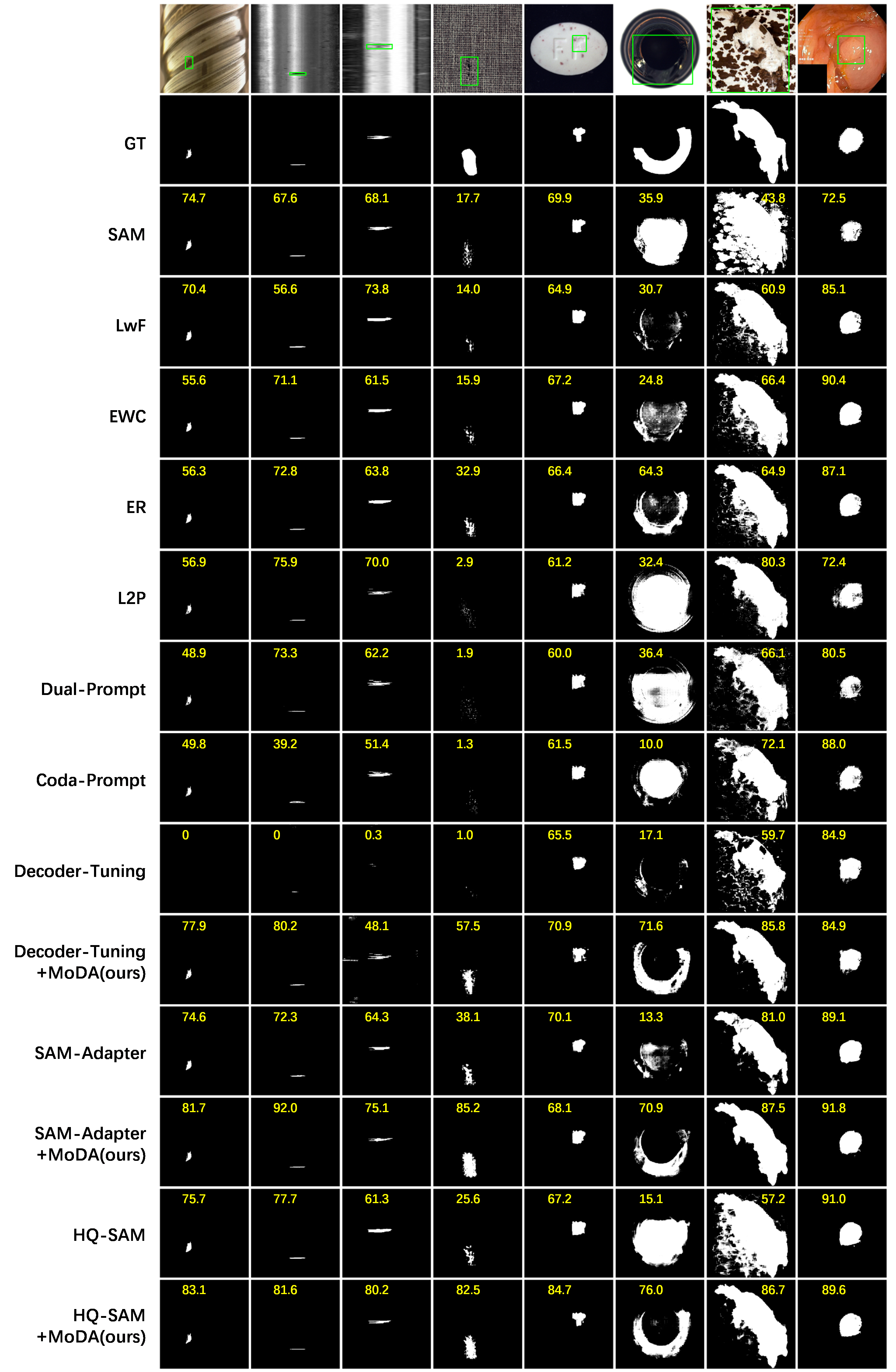}
	\caption{Qualitative comparison of all methods after training on the CoSAM Benchmark}
        \label{fig-sm}
\end{figure*}

\begin{figure*}[t]
	\centering
	\includegraphics[width=0.8\textwidth]{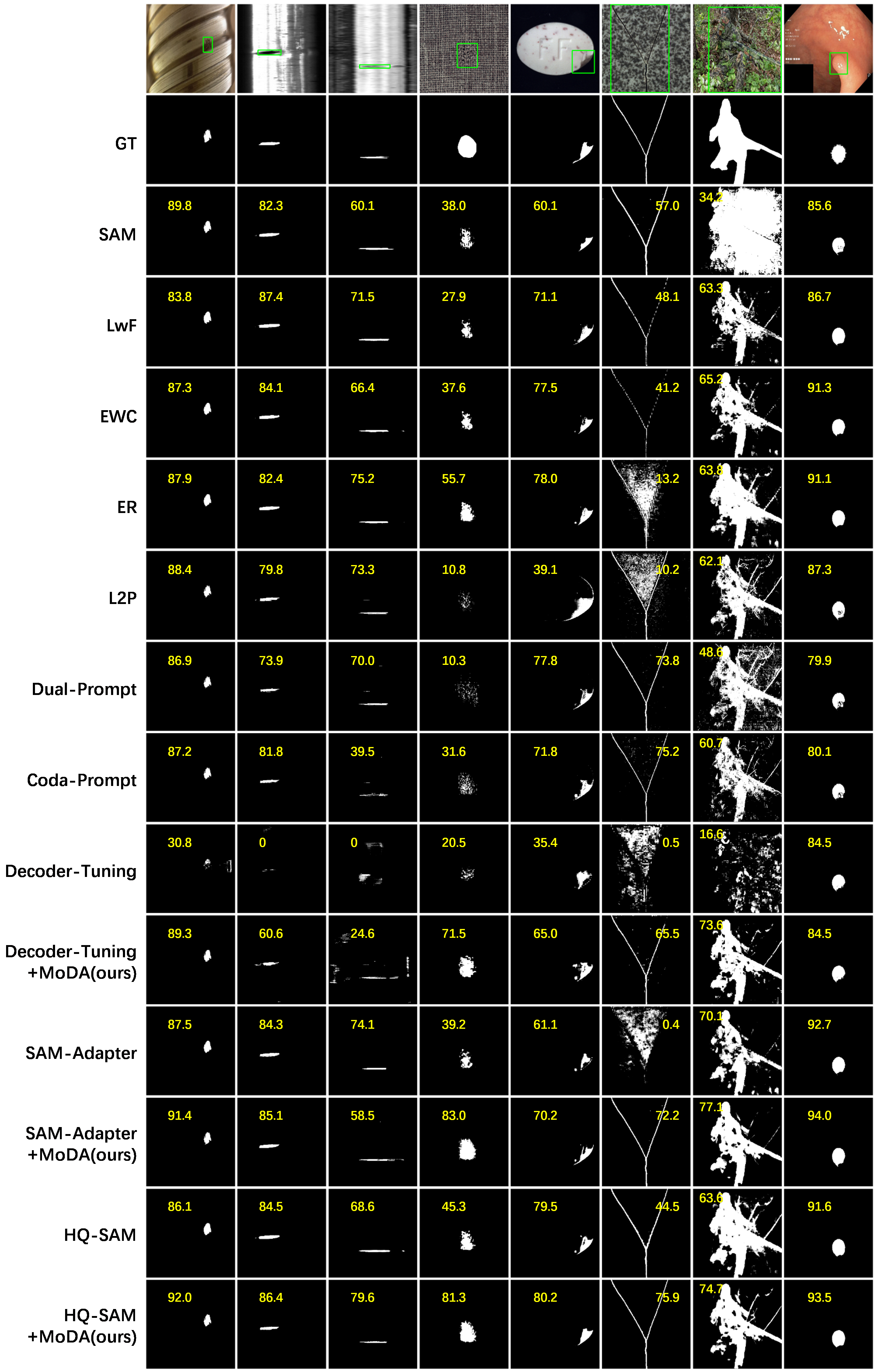}
	\caption{Qualitative comparison of all methods after training on the CoSAM Benchmark}
        \label{fig-sm2}
\end{figure*}

\begin{figure*}[t]
	\centering
	\includegraphics[width=0.8\textwidth]{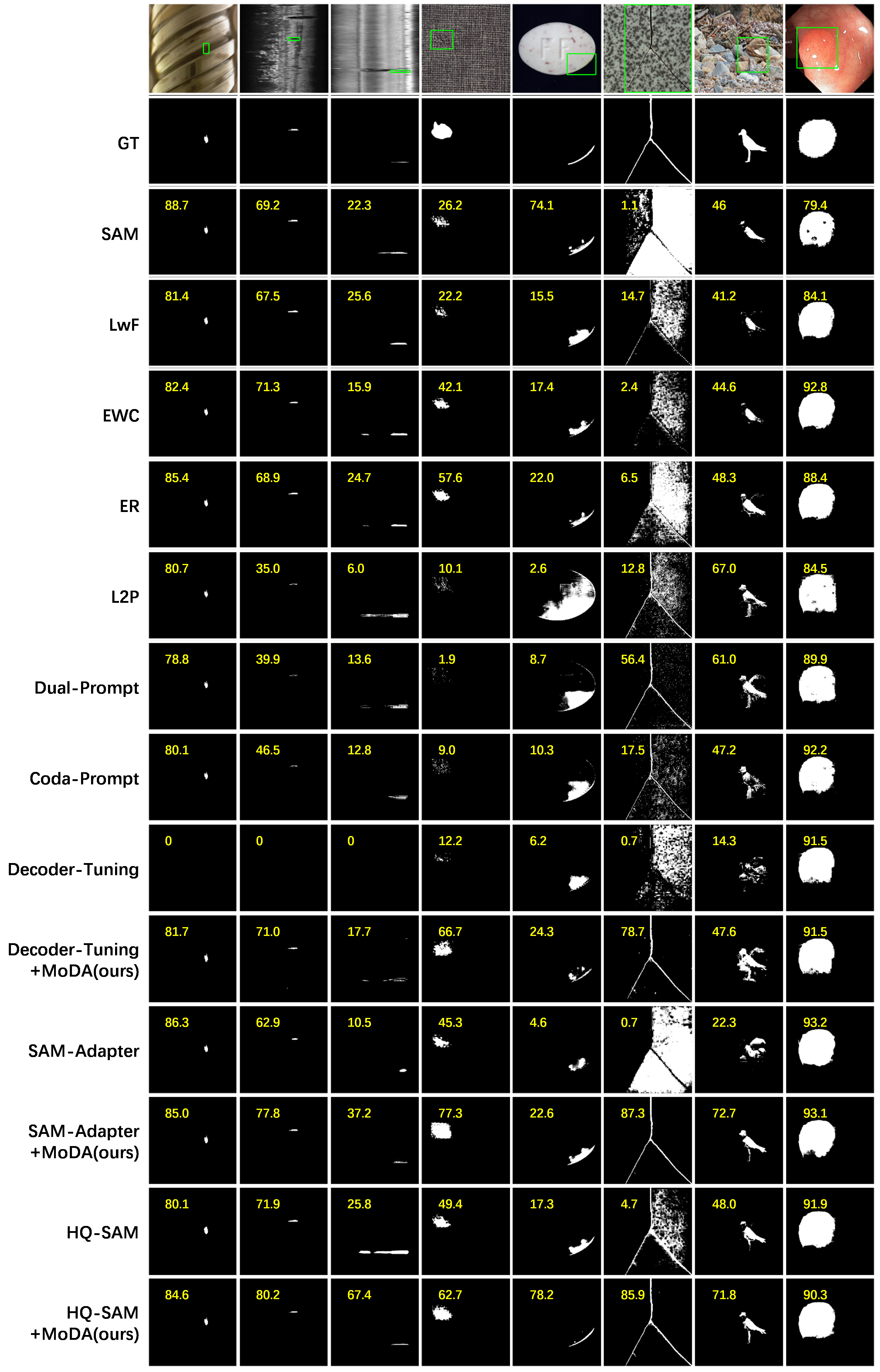}
	\caption{Qualitative comparison of all methods after training on the CoSAM Benchmark}
         \label{fig-sm3}
\end{figure*}

Fig.~\ref{fig-sm} is the complete result of Fig. 4 in the main text, which contains all the comparison methods.
Besides, we provide more visual results on diverse samples after the training on the CoSAM Benchmark, as displayed in Figs.~\ref{fig-sm2} and~\ref{fig-sm3}. Meanwhile, the IoU score is displayed on each mask image for a better comparison. It can be observed that one-step adaptation methods, such as Decoder-Tuning, SAM-Adapter, and HQ-SAM, exhibit severe forgetting in the early tasks, which verifies the significance of the CoSAM Benchmark we proposed that focuses on evaluating the effectiveness of adaptation methods in a continuously changing domain. However, with the help of our proposed MoDA, they consistently achieve obvious performance improvement in all tasks. Especially, for the more difficult Tasks 4, 6, and 7, where the to-be-segmented objects are thin cracks or are highly blended with the background, our method is still capable of obtaining the accurate segmentation mask with clear and sharp edges.

\end{document}